%% file: main_cvpr2023.tex
\documentclass[10pt,twocolumn,letterpaper]{article}

\usepackage{cvpr}              

\usepackage{axessibility}

\usepackage{graphicx}
\usepackage{amsmath}
\usepackage{amssymb}
\usepackage{booktabs}
\usepackage{verbatim}
\usepackage{balance}
\usepackage{multirow}
\usepackage{tabularx}
\usepackage{makecell}
\usepackage[usestackEOL]{stackengine}
\usepackage{mathtools}
\usepackage{wrapfig}

\usepackage{caption}

\usepackage[normalem]{ulem}   
\usepackage[table]{xcolor}
\usepackage{amsmath}

\DeclareMathOperator*{\argmin}{arg\,min}

\usepackage{algorithm}
\usepackage[noend]{algpseudocode}
\algrenewcommand\algorithmicrequire{\textbf{Input:}}
\algrenewcommand\algorithmicensure{\textbf{Output:}}
\algrenewcommand\algorithmicfor{\textbf{For}}
\algrenewcommand\algorithmicwhile{\textbf{While}}

\usepackage{gensymb}

\newcommand{\OursAcronym}{LiDomAug}

\usepackage[pagebackref,breaklinks,colorlinks]{hyperref}

\input{misc/macro}

\def\cvprPaperID{4665} 
\def\confName{CVPR}
\def\confYear{2023}

\begin{document}

\title{Instant Domain Augmentation for LiDAR Semantic Segmentation}

\author{Kwonyoung Ryu$^{1*}$
\hspace{8mm}
Soonmin Hwang$^{2*}$
\hspace{8mm}
Jaesik Park$^1$
\vspace{2mm}\\
$^1$POSTECH 
\hspace{8mm}
$^2$Carnegie Mellon University\\
{\tt \small \url{http://cvlab.postech.ac.kr/research/LiDomAug}}
\vspace{-5mm}
}

\maketitle
\def\thefootnote{*}\footnotetext{Equal contribution}\def\thefootnote{\arabic{footnote}}

\begin{abstract}
\input{section/abstract}
\end{abstract}

\vspace{-3mm}

\input{section/introduction}
\input{section/related_work}
\input{section/method}
\input{section/experiment}
\input{section/conclusion}

\vspace{2mm}
\noindent\textbf{Acknowledgement.}
This work was supported by IITP grant (No.2019-0-01906, POSTECH Artificial Intelligence Graduate School Program) and NRF grant (No.2023R1A1C200781211) funded by the Korean government (MSIT). Hyundai Motor Group provided generous support for this research.

\balance
{\small
\bibliographystyle{ieee_fullname}
\bibliography{egbib}
}

\end{document}


\title{Instant Domain Augmentation for LiDAR Semantic Segmentation \\
\emph{Supplementary Material}}

\author{Kwonyoung Ryu$^{1*}$
\hspace{8mm}
Soonmin Hwang$^{2*}$
\hspace{8mm}
Jaesik Park$^1$
\vspace{2mm}\\
$^1$POSTECH 
\hspace{8mm}
$^2$Carnegie Mellon University\\
{\tt \small \url{http://cvlab.postech.ac.kr/research/LiDomAug}}
}
\maketitle
\def\thefootnote{*}\footnotetext{Equal contribution}\def\thefootnote{\arabic{footnote}}

\renewcommand\thefootnote{\textcolor{red}{\arabic{footnote}}}

This material provides detailed information on handling dynamic objects, label consistency and propagation, detailed LiDAR configurations, and experiment settings. 
We also present per-class domain adaptation results, examples of the generated LiDAR scans, and more qualitative results from 3D semantic segmentation models trained with \OursAcronym.

\section{Dynamic Objects}
\vspace{-2mm}

\begin{algorithm}[h]
    \caption{Dynamic object accumulation}
    \begin{algorithmic}[1]
        \Require{Sequence of point clouds $\{\mathcal{P}_n\}_{n=1}^N$,
        3D bounding boxes $\{\mathbf{b}_n^k\}_{n=1}^N$ of an object $k$, and
        its frame-to-frame transformation matrices $\{\mathbf{T}_n^{k}\}_{n=1}^N$} 
        \Ensure{Accumulated points $\mathcal{P}^k$ of an object $k$}
        \State{$\mathcal{P}^k \longleftarrow \{ \ \}$}
        \State{$\mathbf{T} \longleftarrow \mathbf{I}$}
        \For{$n=1:N$}
        \State{$\mathbf{T} \longleftarrow \mathbf{T}\circ(\mathbf{T}_n^{k})^{-1}$}
        \State{$ \mathcal{P}^k \longleftarrow \{ \mathcal{P}^k \cup \mathbf{T}(\mathcal{P}_n \cap \mathbf{b}_n^k)) \}$}
        \EndFor
    \end{algorithmic}
    \label{alg:dynamic_object_accumulation}
\end{algorithm}

As described in the method section, object-wise motion causes an error, so-called flying points, in constructing the world model if we only use global ego-motion in the aggregation step. If the trajectories of each object over time are provided, the object-wise motion could be canceled out by applying the inverse of them, described in Algorithm \ref{alg:dynamic_object_accumulation}.

As nuScene-lidarseg~\cite{nuscenes_lidarseg} dataset provides the information of object-wise motion in the form of 3D bounding box annotations\footnotemark[1], we apply the Algorithm \ref{alg:dynamic_object_accumulation} to build a better world model.
On the other hand, the object-wise motion information is not available in SemanticKITTI~\cite{semantic_kitti}. In this case, we set a small temporal adjacency for dynamic objects to minimize the error while maximizing the density of the world model, then accumulate the 3D points on dynamic objects by applying global inverse ego-motion.

\footnotetext[1]{We downloaded 3D bounding box annotations from nuScenes full dataset~\cite{nuscenes2019}, not from the nuScenes-lidarseg subset~\cite{nuscenes_lidarseg}.}

\begin{algorithm}[h]
    \caption{Label consistency check and propagation}
    \begin{algorithmic}[1]
        \Require{Point clouds $\{\mathcal{P}_n\}_{n=1}^{N}$}, its semantic label $\{\mathcal{L}_n\}_{n=1}^{N}$, ego-motion $\{\mathbf{T}_n\}_{n=1}^N$, and unlabeled point cloud $\mathcal{P}_u$
        \Ensure{labels $\mathcal{L}$ for $\mathcal{P}_u$}        
        \State{$\mathcal{P}_{world} \longleftarrow \bigcup_{n=1}^N \mathbf{T}_n^{-1}(\mathcal{P}_n)$}        
        \State{$\mathcal{L}_{world} \longleftarrow \bigcup_{n=1}^N \mathcal{L}_n$}
        \State{$\{\mathbf{v}_i\}_{i=1}^{V}\longleftarrow$ Voxelization($\mathcal{P}_{world}$)}
        \For{$i=1 : V$}
        \State{Make $\mathcal{J}=\{j\}$, s.t. $\mathbf{p}_j \in \mathbf{v}_i$ and $\mathbf{p}_j \in \mathcal{P}_{world}$}
        \State{Make $\mathcal{K}=\{k\}$, s.t. $\mathbf{p}_k \in \mathbf{v}_i$ and $\mathbf{p}_k \in \mathcal{P}_{u}$}        
        \State{$l^*_i$ = MostFrequent$(\{l_j|j\in\mathcal{J}\})$, s.t. $l\in\mathcal{L}_{world}$}
        \State{$\mathcal{L}(\mathcal{J})\leftarrow l^*_i$} \Comment{Label consistency}
        \State{$\mathcal{L}(\mathcal{K})\leftarrow l^*_i$} \Comment{Label propagation}        
        \EndFor
    \end{algorithmic}
    \label{alg:label_consistency}
\end{algorithm}

\section{Label consistency check and propagation}

Whereas SemanticKITTI~\cite{semantic_kitti} provides 3D semantic labels for every frame, nuScene-lidarseg~\cite{nuscenes_lidarseg} dataset provides labels only for key-frames sampled at 2Hz. Also, it is difficult for human labelers to label 3D points accurately when the semantics change (e.g., the boundary of the object) or the 3D points are too sparse, so labeling errors occur. As described in the method section, we prepare an examination step for label consistency check and propagation. The 3D points in a world model are divided by a small voxel grid ($0.1m \times 0.1m \times 0.1m$), and the representative labels for each voxel are determined by majority voting. Based on the representative labels, we can correct the semantic labels of 3D points or assign a new label to the unlabeled 3D points. The detailed description of this procedure is shown in Algorithm \ref{alg:label_consistency}. During this procedure, 6.4 points (nuScenes-lidarseg~\cite{nuscenes_lidarseg}) or 8.8 points (SemanticKITTI~\cite{semantic_kitti}) are assigned to a single voxel in average.

\begin{table*}[t]
    \begin{center}    
    \scalebox{0.85}{
        \begin{tabular}{l|c|c|c|c|c}
            \toprule
            \multirow{2}{*}{Model}
            & \multirow{2}{*}{\Centerstack{Velodyne \\ HDL-64E}} 
            & \multirow{2}{*}{\Centerstack{Velodyne \\ HDL-32E}} 
            & \multirow{2}{*}{\Centerstack{Velodyne \\ VLP-16}} 
            & \multirow{2}{*}{\Centerstack{Ouster\\OS-1 64}} 
            & \multirow{2}{*}{\Centerstack{Ouster \\ OS-1 128}} \\ 
            &&&&& \\ \midrule

            {Channels ($H$)}                        & {64}              & {32}                & {16}  & {64}  & {128}              \\ 
            {Range ($r$)}                           & {120 m}           & {100 m}             & {100 m} & {120 m}& {120 m}            \\ 
            {Field of View ($f_{up}, f_{down}$)}        & {+2.0\degree to -24.9\degree } & {+10.67\degree to 30.67\degree } & {+15.0\degree to -15.0\degree } & {+22.5\degree to -22.5\degree } & {+22.5\degree to -22.5\degree } \\ 
            Horizontal angular resolution ($W$) & {2048}  & {2048}    & {2048} & {1024} & {1024} \\
            Clock-wise rotation rate ($\omega_{0}$)    & {20Hz}             & {20Hz}                & {20Hz}  & {20Hz} & {20Hz} \\
            \bottomrule
        \end{tabular}
    }
    \end{center}
    \vspace{-3mm}
    \caption{Detailed configurations of various LiDARs used in our experiments}
    \label{tab:LiDAR_config}
\end{table*}

\begin{figure*}[t]
    \centering
    \includegraphics[width=0.9\textwidth]{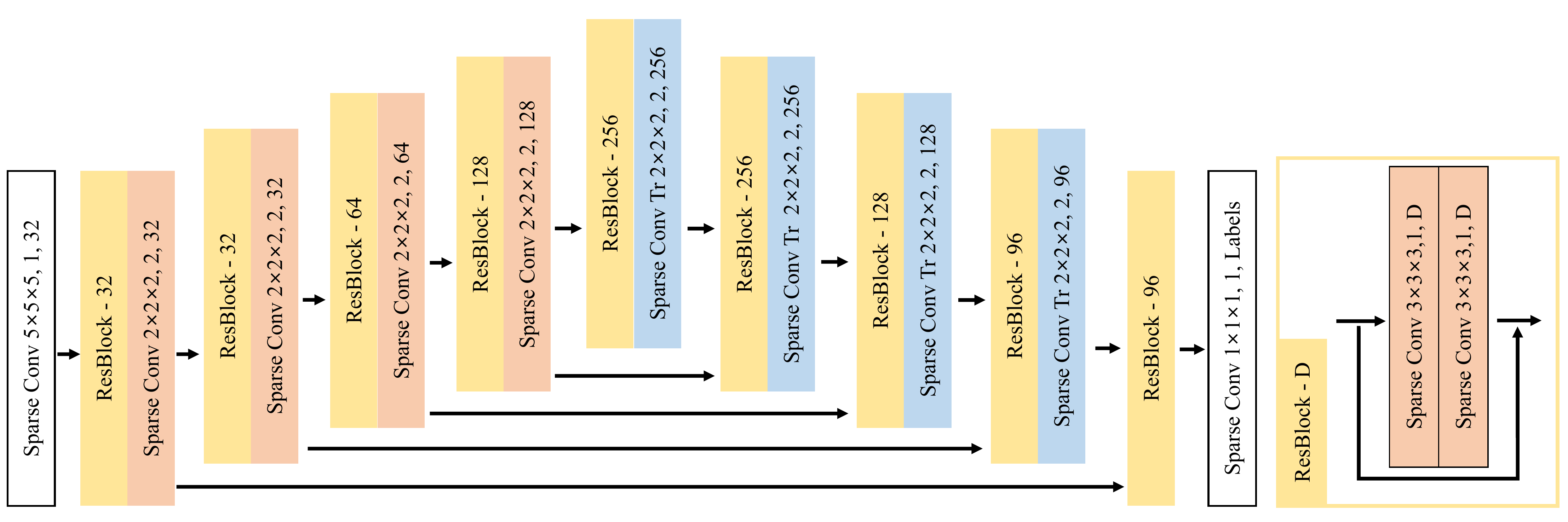}
    \caption{Architecture of MinkowskiNet42(Res16UNet34)\cite{Minkowski}. `Sparse Conv Tr' means transposed sparse convolution.}
    \label{fig:MinkNet42}
\end{figure*}

\vspace{5mm}
\section{Details of LiDAR Configuration}

As mentioned in the experimental section (Table 3), we consider five popular LiDARs from different vendors such as Velodyne\cite{velodyne2007} and Ouster\cite{ouster2018}. Table \ref{tab:LiDAR_config} shows the specifications of the considered LiDARs. 
\vspace{5mm}

\section{Experiment Settings}
\vspace{-6mm}

\noindent \paragraph{Architecture.}
We use MinkNet42~\cite{Minkowski}, and a variant of MinkNet from C\&L~\cite{complete_and_label_cvpr21} and SPVCNN~\cite{SPVNAS_ECCV20} as our baselines. We implement the sparse-convolution-based SPVCNN backbone using MinkowskiEngine~\cite{Minkowski} for a NAS-based backbone experiment. And the MinkNet42 consists of five planes of the encoder and four planes of the decoder. Each res-blocks is set to [32, 64, 128, 256, 256, 128, 96, 96] dimensions with two convolution layers each. Figure \ref{fig:MinkNet42} shows the MinkNet42 architecture we use throughout the experiments. For C\&L backbone, we follow the details in C\&L paper~\cite{complete_and_label_cvpr21} to implement the U-Net style custom MinkNet, which has [24, 32, 48, 64, 80, 96, 112] dimensions for the encoder blocks and [96, 80, 64, 48, 32, 16] dimensions for the decoder blocks. Please refer to the original paper~\cite{complete_and_label_cvpr21} for more details. We use voxel size as $d=5cm$ for all three backbones.
\vspace{2mm}

\noindent \paragraph{Training.}
We use SGD optimizer with the momentum of $0.9$ and weight decay of $10^{-4}$. The initial learning rate is set to $10^{-1}$, decayed by $0.1$ at [3, 8, 15] epochs. We use a batch size of four and use the cross-entropy loss. The training and evaluation are done with a single NVIDIA GeForce RTX 3090 GPU.

\section{Additional Domain Adaptation Scenarios}

\begin{table}[t]
    \newcommand{\minus}[1]{{\textcolor{purple}{$\downarrow$ #1}}}
    \newcommand{\plus}[1]{{\textcolor{teal}{$\uparrow$ #1}}}
    \newcommand{\ccgreen}{\cellcolor{yellow!10}}
    
    \newcommand{\xmark}{\textcolor{teal}{\ding{55}}}%
    \newcommand{\cmark}{\textcolor{purple}{\ding{51}}}%

    \newcolumntype{L}{>{\hspace{0.7em}}l}
    \newcolumntype{R}{>{\hspace{0.6em}}r}

    \centering
    \resizebox{1.0\columnwidth}{!}{
        \setlength{\tabcolsep}{2pt}
        \begin{tabular}{c|L|R@{\hspace{0.5em}}lR@{\hspace{0.5em}}l}    
            \multicolumn{1}{c}{}&\multicolumn{1}{c}{}&\multicolumn{4}{r}{Unit: mIoU (Rel.\%)}\\
            \toprule
            \multicolumn{1}{c|}{\multirow{2}{*}{\begin{tabular}[c]{@{}c@{}}Backbone\\(\# of params)\end{tabular}}}
            &\multicolumn{1}{c|}{\multirow{2}{*}{\begin{tabular}[c]{@{}c@{}}Methods\end{tabular}}}
            &\multicolumn{4}{c}{Source $\rightarrow$ Target} \\
            \cline{3-6}         
            \multicolumn{1}{c|}{} & \multicolumn{1}{c|}{} &\multicolumn{2}{c}{$K \rightarrow P$} & \multicolumn{2}{c}{$N \rightarrow P$} \\                 
            \cmidrule[0.8pt]{1-6}
            \multicolumn{1}{c|}{\multirow{3}{*}{\Centerstack{MinkNet42~\cite{Minkowski}\\(37.8M)}}} 
            & Baseline & \multicolumn{2}{c}{9.5} & \multicolumn{2}{c}{11.3} \\        
            &\ccgreen&\ccgreen&\ccgreen&\ccgreen&\ccgreen\\
            &\multirow{-2}{*}{\ccgreen\shortstack[l]{Baseline\\+ \OursAcronym{}}} & \multirow{-2}{*}{\ccgreen \textbf{11.5}} & \multirow{-2}{*}{\ccgreen(\textbf{\plus{21.1}})}& \multirow{-2}{*}{\ccgreen \textbf{28.1}} & \multirow{-2}{*}{\ccgreen(\textbf{\plus{148.6}})}\\
            
            \bottomrule
        \end{tabular}
    }
    
    \caption{                        
        Training with SemanticKITTI~\cite{kitti} and testing with Pandaset~\cite{pandaset} (K$\rightarrow$P) and vice versa (N$\rightarrow$P) with MinkNet42~\cite{Minkowski}
    }        
    \label{tab:mink_pandaset}
\end{table}

We conduct an additional experiment on \textit{Pandaset}~\cite{pandaset}, which consists of one forward-facing \textit{solid-state LiDAR} and one \textit{spinning Hesai LiDAR}, to investigate whether \OursAcronym{} is effective beyond a single cylindrical LiDAR setup. We train models on SemanticKITTI or nuScenes and evaluate them on \textit{Pandaset}. The results, as shown in Table.~\ref{tab:mink_pandaset}, reveal that our method is effective in both $K\rightarrow P$ and $N\rightarrow P$ scenarios, indicating the potential advantages of our method in complex LiDAR setups.

\begin{table*}[t]
    \newcommand{\ccgreen}{\cellcolor{yellow!10}}
    \resizebox{2.1\columnwidth}{!}{
        \begin{tabular}{c|l|rr|rrrrrrrrrr|r}
        \toprule
        \multicolumn{1}{c|}{Backbone} & Methods & \multicolumn{1}{c}{mIoU} & \multicolumn{1}{c|}{mAcc} &
        \multicolumn{1}{c}{\rotatebox{90}{car}} & 
        \multicolumn{1}{c}{\rotatebox{90}{bicycle}} & 
        \multicolumn{1}{c}{\rotatebox{90}{motorcycle}} & 
        \multicolumn{1}{c}{\rotatebox{90}{truck}} & 
        \multicolumn{1}{c}{\rotatebox{90}{\shortstack{other\\vehicle}}} & 
        \multicolumn{1}{c}{\rotatebox{90}{pedestrian}} & 
        \multicolumn{1}{c}{\rotatebox{90}{\shortstack{drivable\\surface}}} & 
        \multicolumn{1}{c}{\rotatebox{90}{sidewalk}} & 
        \multicolumn{1}{c}{\rotatebox{90}{terrain}} & 
        \multicolumn{1}{c|}{\rotatebox{90}{vegetation}} &
        \multicolumn{1}{c}{\rotatebox{90}{Avg. rank}} \\ \midrule
        \multirow{6}{*}{\rotatebox{90}{MinkNet42}}
        & Baseline & 37.8 & 48.1 & 50.7 & 5.65 & 5.96 & 21.7 & 24.8 & 29.2 & \textbf{89.1} & 42.0 & 23.1 & 85.8 & 3.5\\ 
        \cline{2-15}
        &  CutMix~\cite{Cutmix}  &  37.1 &  46.4 &  75.5 &   0.05 & 14.0 &  26.6 &  22.6 &  3.92 &  86.6 &  36.5 &  19.7 &  85.6 &  4.7\\ 
        &   Copy-Paste~\cite{Copy-Paste} &  38.5 &  48.8 &  77.9 &  3.11 &  11.1 &  21.7 &  31.2 &  7.81 &  88.0 &  38.8 &  19.6 &  86.2 &  3.6\\ 
        &   Mix3D~\cite{Mix3d}   &  43.1 &  52.7 &  72.1 &  0.00 &  34.8 &  11.7 &  26.4 &  28.5 & 83.3 &  41.0 & \textbf{46.4} & \textbf{86.5} & 3.7\\ 
        & Polarmix~\cite{Polarmix} &  45.8 &  54.4 &  74.1&  1.68 &  \textbf{41.9} &  26.9&  23.8 &  \textbf{30.5} &  85.1 &  \textbf{42.7} &  45.3 &  86.2 &  2.7\\ 
        &\ccgreen Baseline+\OursAcronym~ & \ccgreen \textbf{45.9} & \ccgreen \textbf{55.0}& \ccgreen \textbf{79.2}& \ccgreen \textbf{5.78} & \ccgreen 28.0 & \ccgreen \textbf{49.3}& \ccgreen \textbf{32.1} & \ccgreen 13.8 & \ccgreen 88.0 & \ccgreen 42.0 & \ccgreen 35.4 & \ccgreen 85.1 & \ccgreen \textbf{2.4}\\ \bottomrule        
    \end{tabular}}
    \caption{Per-class accuracy from various data augmentation methods. All the models are trained on SemanticKITTI, and tested on nuScene-lidarseg (K$\rightarrow$N).}
    \label{tab:S2N}
\end{table*}

\begin{table*}[!h]
    \newcommand{\ccgreen}{\cellcolor{yellow!10}}
    \resizebox{2.1\columnwidth}{!}{
        \begin{tabular}{c|l|rr|rrrrrrrrrr|r}
        \toprule
        \multicolumn{1}{c|}{Backbone} & Methods & \multicolumn{1}{c}{mIoU} & \multicolumn{1}{c|}{mAcc} &
        \multicolumn{1}{c}{\rotatebox{90}{car}} & 
        \multicolumn{1}{c}{\rotatebox{90}{bicycle}} & 
        \multicolumn{1}{c}{\rotatebox{90}{motorcycle}} & 
        \multicolumn{1}{c}{\rotatebox{90}{truck}} & 
        \multicolumn{1}{c}{\rotatebox{90}{\shortstack{other\\vehicle}}} & 
        \multicolumn{1}{c}{\rotatebox{90}{pedestrian}} & 
        \multicolumn{1}{c}{\rotatebox{90}{\shortstack{drivable\\surface}}} & 
        \multicolumn{1}{c}{\rotatebox{90}{sidewalk}} & 
        \multicolumn{1}{c}{\rotatebox{90}{terrain}} & 
        \multicolumn{1}{c|}{\rotatebox{90}{vegetation}} &
        \multicolumn{1}{c}{\rotatebox{90}{Avg. rank}} \\ \midrule
        \multirow{6}{*}{\rotatebox{90}{MinkNet42}}
        & Baseline & 36.1 & 46.9 & 78.5 & 0.00 & 8.17 & 3.37 & {11.1} & \textbf{34.5} & 66.3 & 35.8 & 39.4 & {84.2} & 4.5\\ 
        \cline{2-15}
        &  CutMix~\cite{Cutmix}  &  37.6 &  51.4 &  81.2 &   0.00 &  5.28 &  9.09 &  \textbf{17.4} &  11.8 &  73.6 &  45.5 &  46.8 &  85.7 &  3.9\\ 
        &   Copy-Paste~\cite{Copy-Paste} &  41.1 &  56.6 &  85.7 &  0.00 &  8.17 &  12.8 &  6.46 &  28.6 &  \textbf{80.8} &  \textbf{47.4} &  53.8 &  87.2 &  2.8\\ 
        &   Mix3D~\cite{Mix3d}   &  44.7 &  63.9 &  \textbf{93.1} &  10.4 &  31.3 &  17.0 &  14.1 &  {34.2} &  {71.8} &  40.7 &  {44.6}&  \textbf{89.5} &  {2.8}\\ 
        & Polarmix~\cite{Polarmix} &  {39.1} &  {61.2}&  {75.9}&  19.4 &  19.7 &  {9.60}&  {3.03} &  18.3 &  75.0 &  {43.1} &  48.9 &  77.8 &  {4.0}\\ 
        &\ccgreen Baseline+\OursAcronym~ & \ccgreen \textbf{48.3} & \ccgreen \textbf{69.0}& \ccgreen 92.6& \ccgreen \textbf{31.6} & \ccgreen \textbf{42.5} & \ccgreen \textbf{21.6}& \ccgreen 6.43 & \ccgreen 34.4 & \ccgreen 70.0 & \ccgreen 47.1 & \ccgreen \textbf{59.4} & \ccgreen 77.5 & \ccgreen \textbf{2.6}\\ \bottomrule
    \end{tabular}}
    \caption{Per-class accuracy from various data augmentation methods. All the models are trained on nuScene-lidarseg, and tested on SemanticKITTI (N$\rightarrow$K).
    \label{tab:N2S}
    }
\end{table*}

\section{Per-class Domain Adaptation Results}

In Table \ref{tab:S2N} (K$\rightarrow$N) and Table \ref{tab:N2S} (N$\rightarrow$K), we present per-class IoU numbers from our experiments, and Baseline + \OursAcronym{} achieves the best performances in mIoU, mAcc, and avg. rank metrics. Particularly, as shown in Table \ref{tab:N2S} (N$\rightarrow$K), our Baseline + \OursAcronym{} setting, in which the label propagation and dynamic object handling are applied, significantly improves the performance of moving object classes (bicycle: $0.00 \rightarrow 31.6$, motorcycle: $8.17 \rightarrow 42.5$, truck: $3.37 \rightarrow 21.6$). Note that our final model, Baseline+\OursAcronym{}, achieves the best performance in average rank across classes, implying that \OursAcronym{} helps the model generalize over multiple classes beyond sensor-shift.
 
\newpage

\section{More Qualitative Results}

\begin{figure*}[b]
    \centering
    \includegraphics[width=1\textwidth]{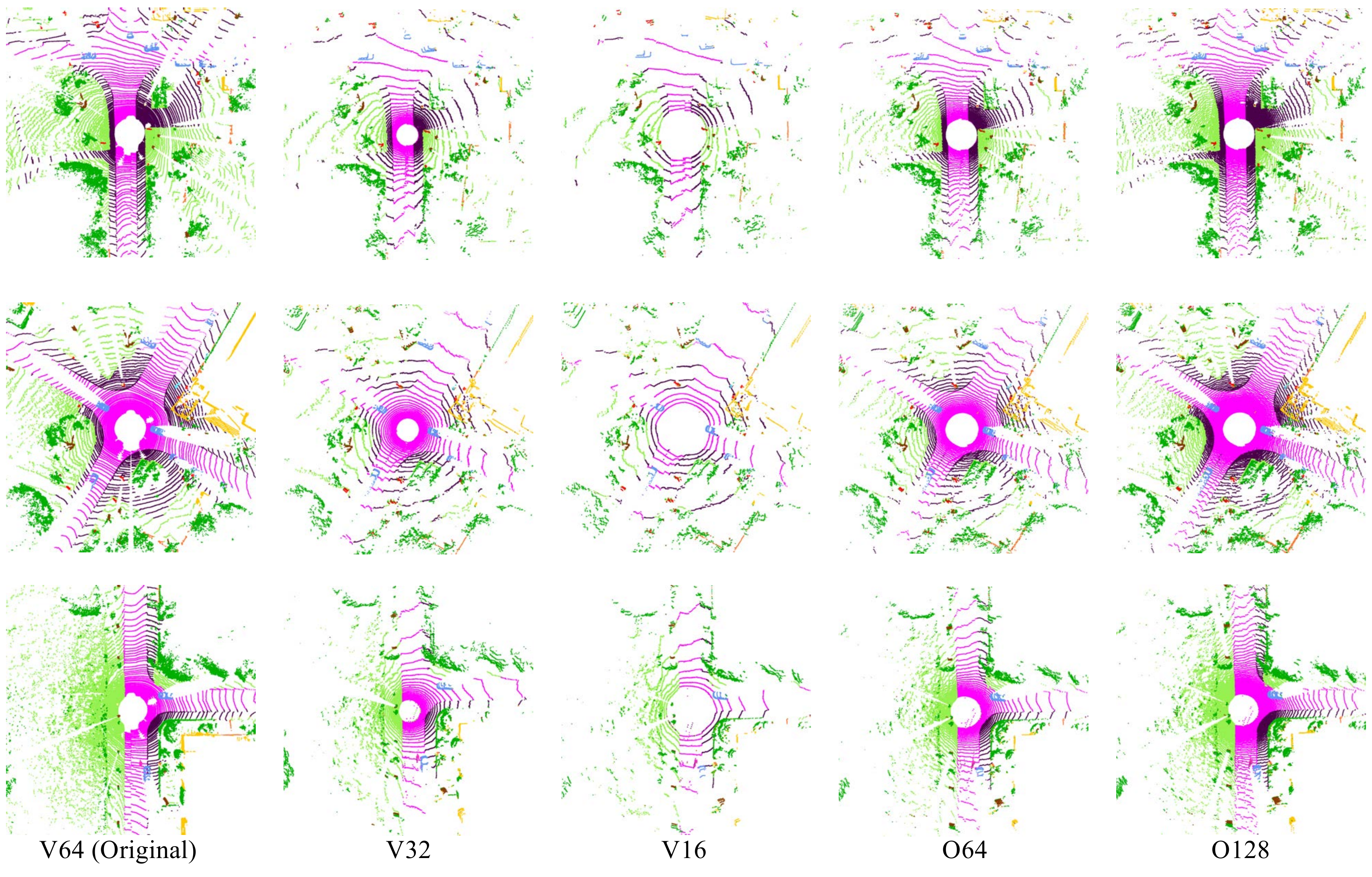}

    \caption{Augmented LiDAR frames using SemanticKITTI dataset. We augment 5 types of LiDARs (Table~\ref{tab:LiDAR_config}). We mark V64 as the original frame because SemanticKITTI is captured with Velodyne HDL-64E.}
    \label{fig:simulation}
\end{figure*}

\begin{figure*}[t]
    \centering
    \includegraphics[width=1\textwidth]{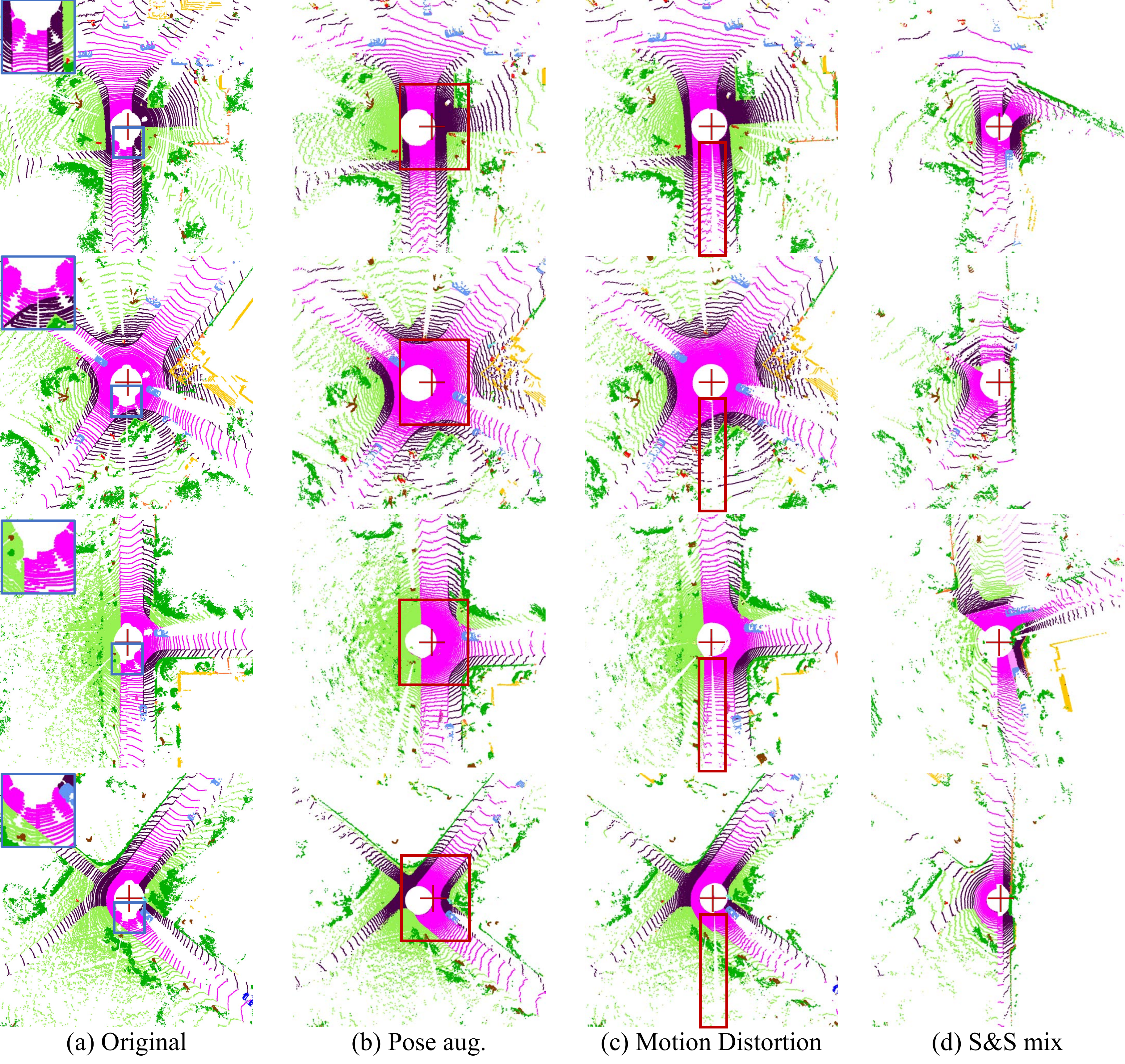}

    \caption{Data augmentation using \OursAcronym on SemanticKITTI.  (a) Original frames (b) Generated frames after pose augmentation (c) Rendered frames with the consideration of frame distortion (d) Mixed two scenes with various LiDAR configurations. \textcolor{red!80}{The red cross} means the center of the original frames, and \textcolor{brightmaroon}{the dark red box} highlights the changes. Note that the motion distortion exists in the original frames(1st,3rd row: distorted by LiDAR move straight, 2nd row : distorted by LiDAR rotate counter clock-wise, 4th row : distorted by LiDAR rotate counter clock-wise) as \textcolor{blue!80}{the blue box} highlighted.}
    \label{fig:augmentation}
\end{figure*}

\begin{figure*}[t]
    \centering
    \includegraphics[width=1\textwidth]{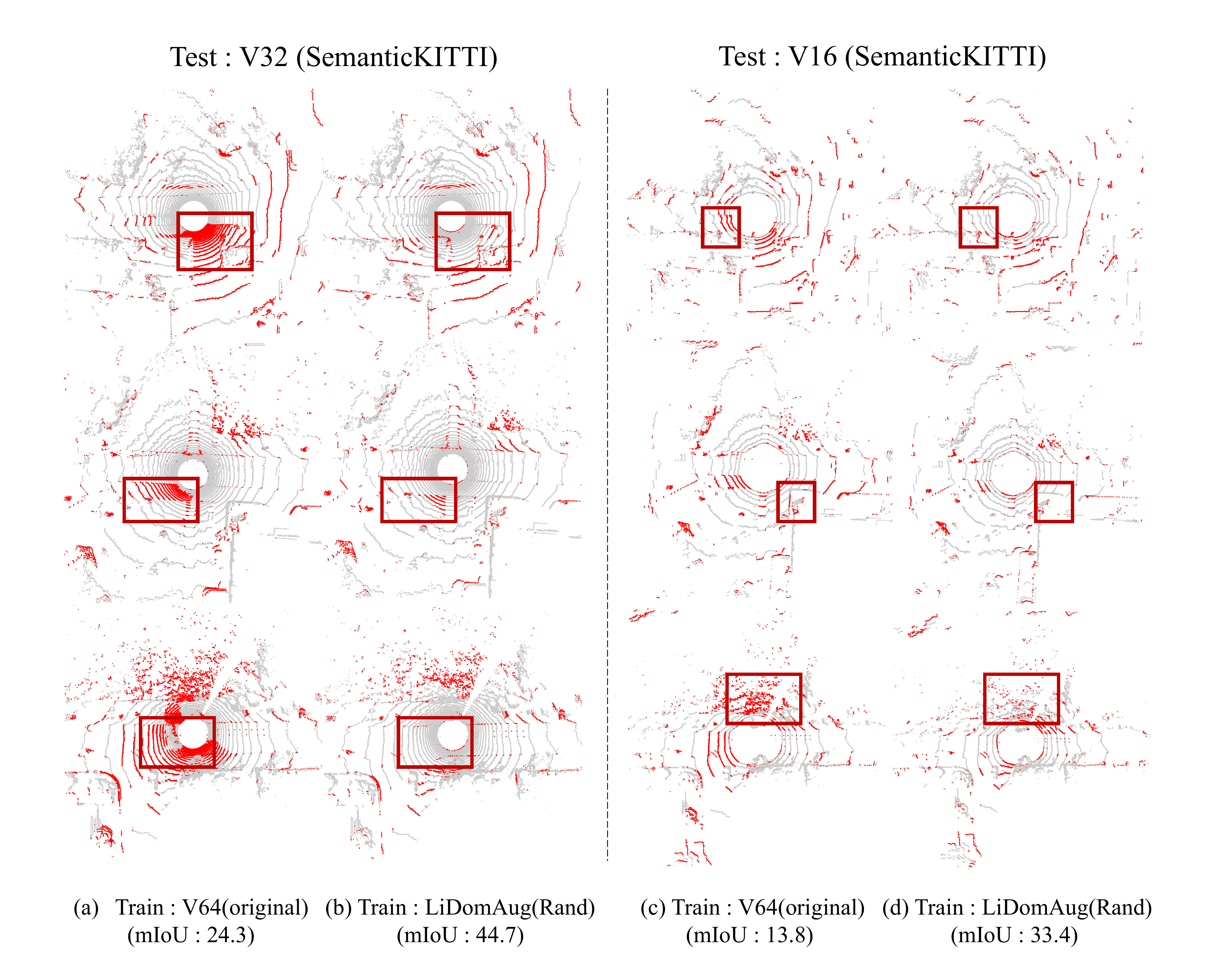}
    
    \caption{Evaluation of inconsistent LiDAR configuration (corresponds to Table 3 in the main paper). All the models here are trained with/without \OursAcronym{} on SemanticKITTI. \textcolor{red!80}{The red points} indicate the wrong predictions. Train$\rightarrow$Test: (a) V64 (original; w/o \OursAcronym{})$\rightarrow$V32 (b) \OursAcronym{}$\rightarrow$V32 (c) V64 (original; w/o \OursAcronym{})$\rightarrow$V16 (d) \OursAcronym{} $\rightarrow$V16. \textcolor{brightmaroon}{The dark red box} highlights the changes. As shown in columns (b) and (d), \OursAcronym{} helps relieve erroneous predictions from inconsistent LiDAR types in training and testing.}
    \label{fig:k2k_comparison_qual}
\end{figure*}

\begin{figure*}[t]
    \centering
    \includegraphics[width=1\textwidth]{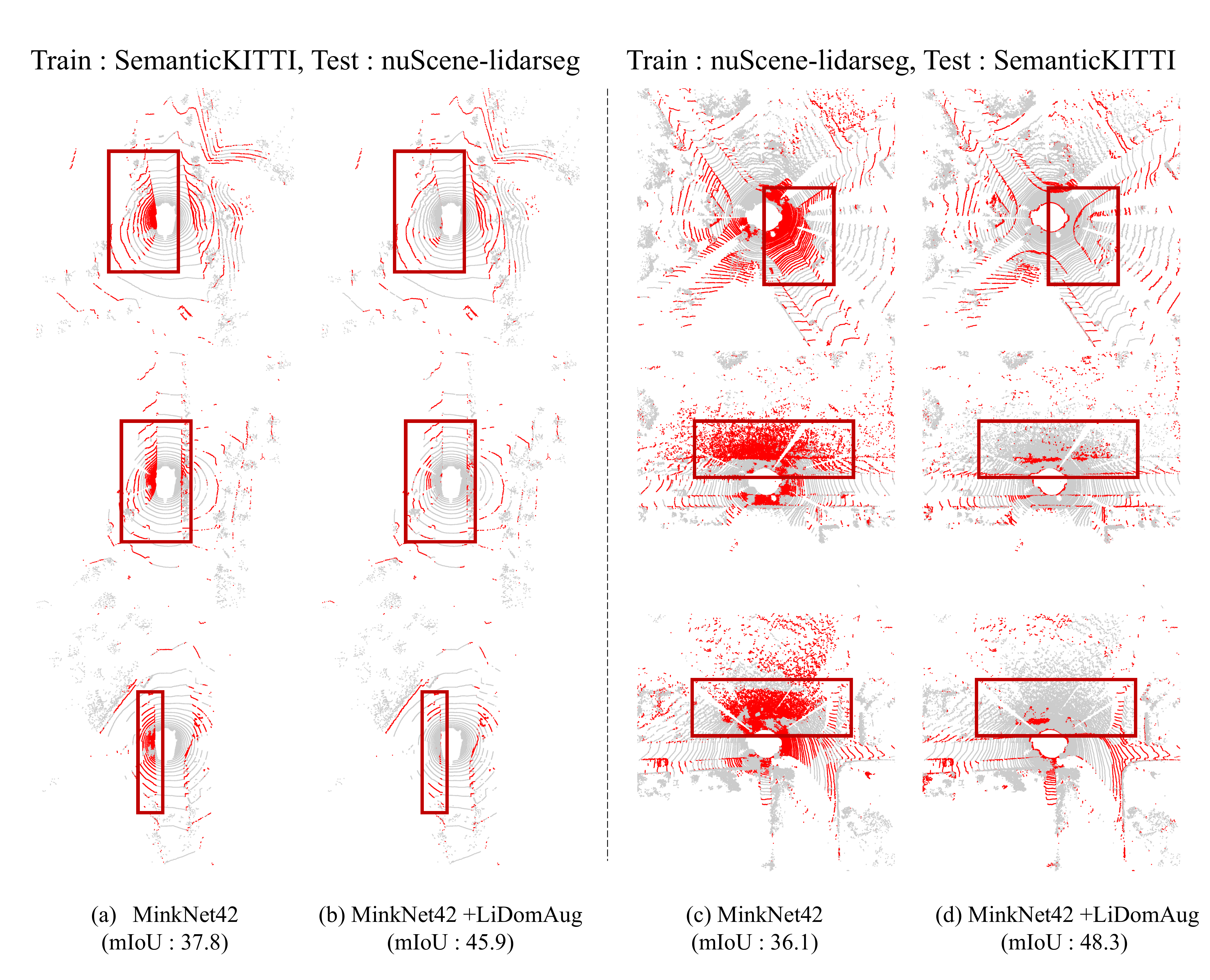}
    
    \caption{Evaluation on domain adaptation settings with MinkNet42~\cite{Minkowski} backbone (corresponds to Table 1 in main paper). \textcolor{red!80}{The red points} indicate the wrong predictions. Train$\rightarrow$Test: (a) Baseline MinkNet42 (source LiDAR setting V64 $\rightarrow$ target LiDAR setting V32) (b) Trained with our \OursAcronym{} (c) Baseline MinkNet42 (source LiDAR setting V32 $\rightarrow$ target LiDAR setting V64) (d) Trained with our \OursAcronym{}. \textcolor{brightmaroon}{The dark red box} highlights the changes. \OursAcronym{} (b, d) helps to relieve erroneous predictions from dataset shifts in training and testing.}
    \label{fig:k2n_n2k_comparison_qual}
\end{figure*}

\begin{figure*}[t]    
    \centering
    \includegraphics[width=1\textwidth]{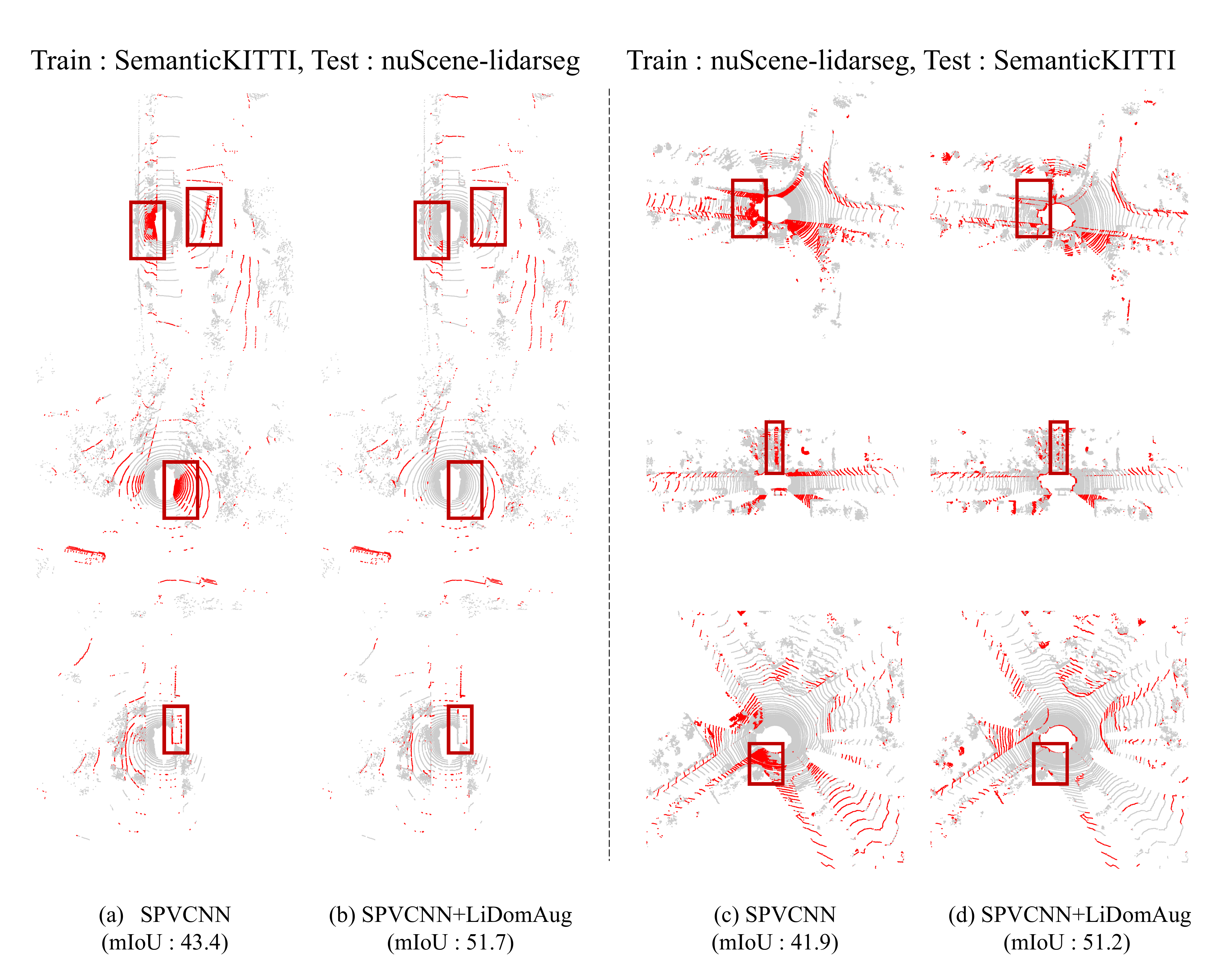}
    
    \caption{Evaluation on domain adaptation settings with SPVCNN~\cite{SPVNAS_ECCV20} backbone (corresponds to Table 5 in main paper). \textcolor{red!80}{The red points} indicate the wrong predictions. Train$\rightarrow$Test: (a) Baseline SPVCNN (source LiDAR setting V64 $\rightarrow$ target LiDAR setting V32) (b) Trained with our 
    \OursAcronym{} (c) Baseline SPVCNN (source LiDAR setting V32 $\rightarrow$ target LiDAR setting V64) (d) Trained with our 
    \OursAcronym{}. \textcolor{brightmaroon}{The dark red box} highlights the changes. \OursAcronym{} (b, d) helps to relieve erroneous predictions from dataset shifts in training and testing.}    
    \label{fig:k2n_n2k_comparison_qual_spvcnn}
\end{figure*}

We included a video file, LiDomAug.mp4, to showcase (1) augmented frames using \OursAcronym{}, (2) dynamic object accumulation, and (3) label consistency check and propagation. We also present the qualitative results of (4) the evaluation of generated LiDAR frames and (5) the evaluation of two different datasets. Figure \ref{fig:simulation} and \ref{fig:augmentation} represent the qualitative examples from \OursAcronym{} in SemanticKITTI. Our data generation can imitate the LiDAR geometric patterns with various configurations. Figure \ref{fig:k2k_comparison_qual}, \ref{fig:k2n_n2k_comparison_qual}, \ref{fig:k2n_n2k_comparison_qual_spvcnn} shows the qualitative comparison between the model trained with and without \OursAcronym{}, and show significant improvement.

\clearpage
\balance
{\small
\bibliographystyle{ieee_fullname}
\bibliography{egbib}
}

%% file: misc/macro.tex
\usepackage[capitalize]{cleveref}
\crefname{section}{Sec.}{Secs.}
\Crefname{section}{Section}{Sections}
\Crefname{table}{Table}{Tables}
\crefname{table}{Tab.}{Tabs.}

\renewcommand\thefootnote{\textcolor{red}{\arabic{footnote}}}

\usepackage{pifont}

%% file: section/abstract.tex
Despite the increasing popularity of LiDAR sensors, perception algorithms using 3D LiDAR data struggle with the \textbf{sensor-bias problem}. Specifically, the performance of perception algorithms significantly drops when an unseen specification of the LiDAR sensor is applied at test time due to the domain discrepancy.
This paper presents a fast and flexible LiDAR augmentation method for the semantic segmentation task called \textbf{\OursAcronym}. It aggregates raw LiDAR scans and creates a LiDAR scan of any configurations with the consideration of dynamic distortion and occlusion, resulting in instant domain augmentation. Our on-demand augmentation module runs at \textbf{330 FPS}, so it can be seamlessly integrated into the data loader in the learning framework.
In our experiments, learning-based approaches aided with the proposed \OursAcronym{} are less affected by the sensor-bias issue and achieve new state-of-the-art domain adaptation performances on SemanticKITTI and nuScenes dataset without the use of the target domain data. We also present a sensor-agnostic model that faithfully works on the various LiDAR configurations.

%% file: section/introduction.tex
\section{Introduction}
\label{sec:introduction}

LiDAR (Light Detection And Ranging) is a modern sensor that provides reliable range measurements of environments sampled from 3D worlds and has become crucial for intelligent systems such as robots~\cite{behley2018rss,LOAM}, drones~\cite{drone}, or autonomous vehicles~\cite{LiDAR_autonomous_article, kitti}. Therefore, developing resilient 3D perception algorithms for LiDAR data~\cite{RandLANet_CVPR20,PointPillars,PseudoLiDAR} is becoming more crucial.

\begin{figure}[t]
    \begin{center}
        \includegraphics[width=1.0\linewidth]{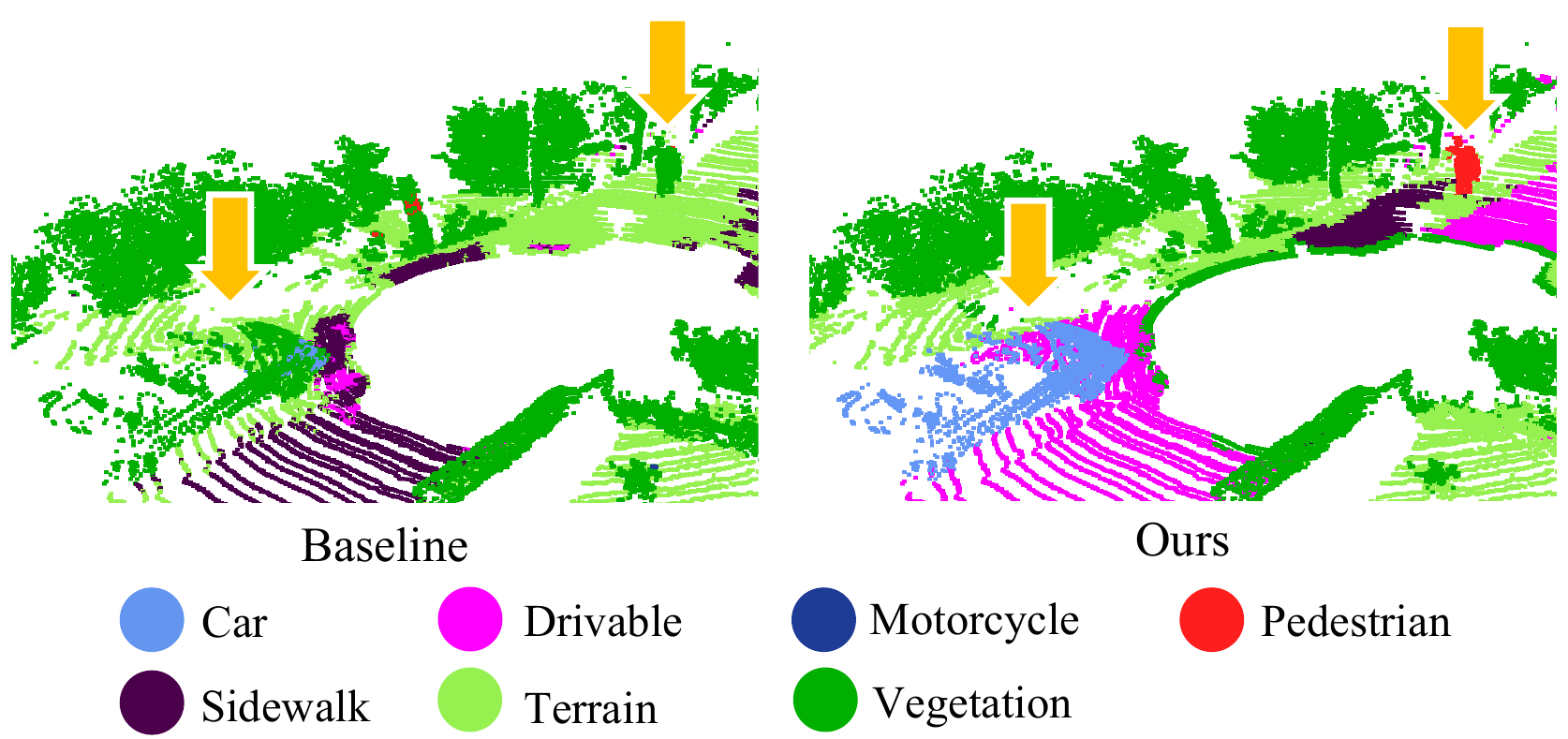}
    \end{center}
    \vspace{-3mm}
    
    \caption{
        Data-driven LiDAR semantic segmentation methods often show an accuracy drop when they are applied to unseen LiDAR configurations. (Left) A result of a baseline approach, where inaccurate predictions are highlighted with orange arrows. (Right) A result of the baseline approach aided with the proposed~\OursAcronym{}. For both results, we use nuScenes~\cite{nuscenes2019} (32 ch.) for the training and use SemanticKITTI~\cite{semantic_kitti} (64 ch.) for the testing.        
    }
    \vspace{-3mm}
    \label{teaser_introduction} 
\end{figure}

With the growing interest in LiDAR sensors, various LiDAR sensors from multiple manufacturers have become prevalent. As a result, popular 3D datasets~\cite{nuscenes2019,chang2019argoverse,kitti,geyer2020a2d2,huang2018apolloscape} are captured by different LiDAR configurations, which are defined by vertical/horizontal resolutions, a field of view, and a mounting pose. Due to the difference in sampling patterns from various LiDAR configurations, the \emph{sensor-bias problem} arises in 3D perception algorithms~\cite{complete_and_label_cvpr21,triess2021survey}. For example, as shown in Figure~\ref{teaser_introduction}, we observe a severe performance drop in LiDAR semantic segmentation task if the LiDAR used to collect the test set differs from the LiDAR used for the training set.

Although the sensor-bias problem is crucial, an existing solution, such as domain adaptation, is tuned for a specific LiDAR configuration, which is suboptimal to designing a sensible 3D perception method. Specifically, Supervised Domain Adaptation requires massive labeling costs to learn to adapt to the new data captured with a target sensor. Hence, such an approach is often not viable in practice. Unsupervised Domain Adaptation~\cite{jaritz2020xmuda,jiang2021lidarnet,complete_and_label_cvpr21} aims to make a model adapt to a target domain without using direct annotations. However, there is an accuracy degradation, and such approaches require enough collection of target domain data. Thus, it is demanding to design a new approach that can be applied instantly to an unseen target domain without requesting any target domain data.

By focusing on the widely used cylindrical LiDARs, this paper presents a new approach to alleviate the sensor-bias problem. The proposed method, called \emph{\OursAcronym{}}, augments the training data based on arbitrary cylindrical LiDAR configurations, mounting pose, and motion distortions. The proposed on-demand augmentation module runs at 330 FPS, which can be regarded as an \emph{instant domain augmentation}. This flexibility, which is a key strength of our method, enables us to train a sensor-agnostic model that can be directly applied to multiple target domains.

We demonstrate our method on the task of LiDAR semantic segmentation. In particular, we tackle the domain discrepancy problem when the LiDAR sensors used for making the training and the test data are not consistent. Interestingly, learning-based approaches aided with the proposed \OursAcronym{} outperform the state-of-the-art Unsupervised Domain Adaptation approaches~\cite{FeaDA,SWD,OutDA,3DGCA,complete_and_label_cvpr21} without access to any target domain data. Our method also beats Domain Mapping~\cite{domain_transfer_iros20, bevsic2021unsupervised} and Domain Augmentation approaches~\cite{Cutmix,Copy-Paste, Mix3d, Polarmix}, showing the practicality of the proposed approach. In addition, we show a semantic segmentation model trained with \OursAcronym{} that works faithfully on the various cylindrical LiDAR configurations.

Our contributions can be summarized as follows:
\begin{itemize}
    \renewcommand{\labelitemi}{$\bullet$}
    \item We present an \textit{instant LiDAR domain augmentation} method, called \OursAcronym{}, for LiDAR semantic segmentation task. Our on-demand augmentation module runs at 330 FPS. 
    \item Our method can augment arbitrary cylindrical LiDAR configurations, mounting pose, and entangled motions of LiDAR spin and moving platform just from the input data. We empirically validate that such flexible modules are helpful in learning sensor-agnostic LiDAR frameworks.
    \item Experiments show that LiDAR semantic segmentation networks trained with the proposed \OursAcronym{} outperform the state-of-the-art Unsupervised Domain Adaptation, Domain Mapping, and LiDAR Data Augmentation approaches.     
\end{itemize}

%% file: section/related_work.tex
\begin{figure*}[ht!]
    \begin{center}
        \includegraphics[width=1\linewidth]{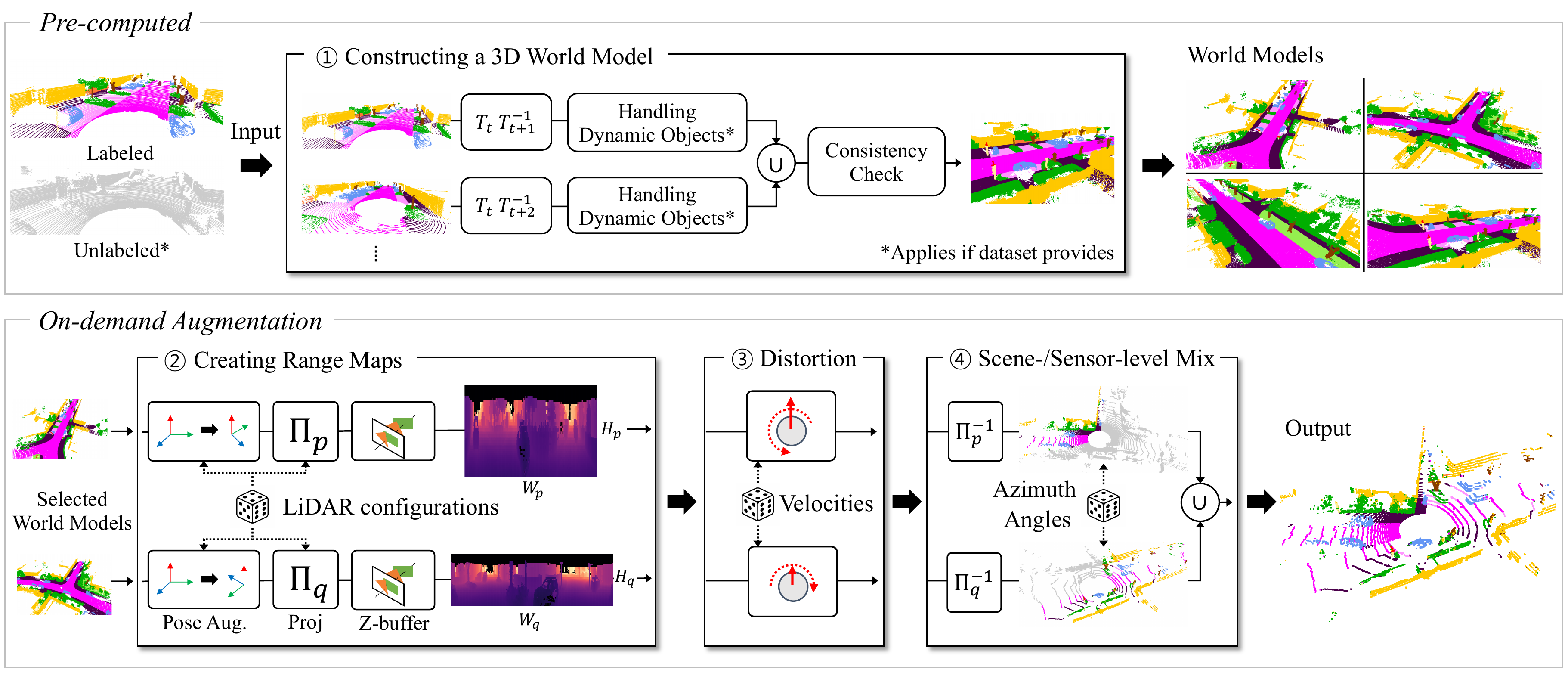}
    \end{center}
    \vspace{-12pt}
    
    \caption{
        \textbf{Overview of the proposed \OursAcronym{} pipeline.} 
        \ding{192} A dense 3D world model is constructed from raw LiDAR frames with the consideration of unlabeled frames and dynamic objects (Sec.~\ref{sec:construct_a_world_model}). \ding{193} Range maps of arbitrary LiDAR configuration are rendered by projecting the world model after applying random pose augmentation (Sec.~\ref{subsec:sensor-agnostic LiDAR projection}). \ding{194} The distortion induced by LiDAR spin and ego-motion is applied (Sec.~\ref{subsec:Apply Situation-aware Transformation}). \ding{195} Range maps are mixed using random azimuth ranges and back-projected to make an output 3D point cloud (Sec.~\ref{sec:sensor_level_mix}). Note that the on-demand augmentation module that comprises \ding{193}, \ding{194}, and \ding{195} runs at 330FPS.
    }        
    \label{fig:overview}        
\end{figure*}

\section{Related Work}
\label{sec:related work}

\subsection{LiDAR Domain Adaptation and Mapping}
\textbf{Domain Adaptation.}
A representative direction to alleviate the sensor-to-sensor domain shift issue is to adopt domain adaptation approaches~\cite{triess2021survey}. Cross-modal learning~\cite{jaritz2020xmuda} is exploited to enable controlled information exchange between image predictions and 3D scans. Adversarial domain adaptation methods are introduced for output space~\cite{OutDA} or feature space alignment~\cite{FeaDA} by employing sliced Wasserstein discrepancy~\cite{SWD} or boundary penalty~\cite{jiang2021lidarnet}. 3DGCA\cite{3DGCA} aligns the statistics between batches from source and target data with geodesic distance. A sparse voxel completion network~\cite{complete_and_label_cvpr21} is proposed to learn a mapping from the source domain to a canonical domain that contains complete and high-resolution point clouds. LiDAR semantic segmentation is performed on the canonical domain, and the result is projected to the target domain. ConDA\cite{ConDA} and CoSMix\cite{CoSmix} also construct an intermediate domain by mixing or concatenating the source and the target domains using pseudo-labeled target data to mitigate the domain shift issue. GIPSO\cite{Saltori2022GIPSOGI}, a recent online adaptation method, requires an optimization process on target domain data using geometric propagation and temporal regularization with pseudo labels inferred from a source domain model. A common limitation of the above methods is to require additional optimization with access to target domain data, which hinders their practicality. On the other hand, our method only adds slight augmentation overhead in the training phase and circumvents the need for target domain data.
\vspace{2mm}

\noindent\textbf{Domain Mapping.}
Our most relevant approach is domain mapping that directly transforms the source domain data to the target-like LiDAR scan~\cite{bevsic2021unsupervised,domain_transfer_iros20} and uses the transformed data for the training. However, the approach by Bešić~\etal~\cite{bevsic2021unsupervised} requires access to target domain data, and the method proposed by Langer~\etal~\cite{domain_transfer_iros20} is computationally heavy due to mesh operations that recover surfaces and check occlusions. Instead, our method can produce various LiDAR scans considering multiple LiDAR configurations in 330 FPS. Our experiment shows the efficacy of our LiDAR scans on the LiDAR semantic segmentation task.

\subsection{LiDAR Data Augmentation}

Approaches for LiDAR data augmentation have been explored in various ways. Inspired by seminal work in image augmentation~\cite{Cutmix}, augmentation methods for the LiDAR object detection task~\cite{Augmented_RA_L_2020, Cheng2020Improving3O,Choi2021PartAwareDA,LiDARAug_CVPR2021,Pattern-aware, Vfield} are proposed. However, these works are crafted for the detection task and assume bounding box labels are provided. For the 3D semantic segmentation task, CutMix~\cite{Cutmix} and Copy-Paste~\cite{Copy-Paste} extend the successful ideas applied for 2D image augmentation. Mix3D~\cite{Mix3d} aggregates the two 3D scenes to make objects implicitly placed into a novel out-of-context environment, which encourages the model to focus more on the local structure. Recently, PolarMix~\cite{Polarmix} introduces the scene- and object-level mix in cylindrical coordinates. PolarMix shows an impressive performance gain in domain adaptation tasks but is limited to demonstrating its synthetic-to-real adaptation capability. To the best of our knowledge, our approach is the first work on comprehensive LiDAR data augmentation to address the sensor-bias issue, and it shows superior performance compared with existing 3D data augmentation approaches.

\subsection{LiDAR Semantic Segmentation}

Existing approaches for 3D semantic segmentation can be categorized into three groups: 2D projection-based, point-based, and voxel-based methods. The 2D projection-based approaches~\cite{kochanov2020kprnet,milioto2019rangenet++,xu2020squeezesegv3} project 3D point clouds to 2D space and apply a neural network architecture crafted for image perception. Point-based methods directly work on unstructured and scattered point cloud data. Approaches in this category utilize point-wise multi-layered perceptron~\cite{Pointnet,Pointnet++}, point convolution~\cite{PointCNN,liu2019pvcnn,KPConv}, or lattice convolution~\cite{Splatnet}. Voxel-based methods handle voxelized 3D points. Early work~\cite{Volumetric_CVPR16,3dshapenet_CVPR15} adopts dense 3D convolutions, but recent approach~\cite{Minkowski} regards voxel as a sparse tensor and presents an efficient semantic segmentation framework.

Among these approaches, we select KPConv~\cite{KPConv} (point-based) and MinkowskiNet~\cite{Minkowski} (voxel-based) for the semantic segmentation experiments due to their efficiency and fidelity in the field. We apply the proposed \OursAcronym{} to the selected networks to see the improvements.

%% file: section/method.tex
\section{Fast LiDAR Data Augmentation} 
\label{sec:method}

We introduce a new augmentation method, called \textbf{\OursAcronym{}}, that instantly creates a new LiDAR frame considering LiDAR mounting positions, various LiDAR configurations, and distortion caused by LiDAR spin and ego-motion. In this work, we craft our augmentation approach for cylindrical LiDARs. As shown in Fig.~\ref{fig:overview}, our method consists of four steps: \ding{192} Constructing a world model from LiDAR frames, \ding{193} Creating a range map of arbitrary LiDAR configurations and poses, \ding{194} Applying motion distortion to the augmented frames caused by ego-motion, and \ding{195} Scene-level \& sensor-level mix. The proposed method is flexible enough to produce a combined LiDAR frame having multiple LiDAR configurations.

\subsection{Constructing a 3D World Model}
\label{sec:construct_a_world_model}

A LiDAR frame is partial geometric capture of a 3D world. Therefore, we can aggregate multiple LiDAR frames of similar regions to build a rough 3D world model. In this step, we separately care for static scenes and dynamic objects by utilizing semantic label annotations on 3D points and trajectories of moving objects in the scene. Such information is available in standard LiDAR datasets~\cite{nuscenes_lidarseg}.
\vspace{2mm}

\noindent\textbf{Static scene.}
 We construct a static world model by aggregating multiple LiDAR frames using ego-motion. Specifically, a set of motion-compensated LiDAR frames $\mathcal{P}_t ^{world}$ is built as a world model at time $t$ by aggregating $N$ adjacent LiDAR frames. We determine the adjacent LiDAR frames using geometric adjacency (based on the LiDAR center coordinates) rather than temporal adjacency (based on frame indices) to cover the 3D scene better. This scheme helps to build a denser world map when the ego vehicle revisits the same place, formulated as follows:
\begin{equation}
        \mathcal{P}_t^{world} =\bigcup_{k\in {K}_{t}}{T_{t}\circ T_{t+k}^{-1}(\mathcal{P}_{t+k})},        
    \label{equ:world_model_from_static_scene}    
\end{equation}
where the geometrically adjacent set of frames ${K}_{t} = \argmin_{K} \sum_{k\in K}{||\mathbf{R}_{t+k}^\top \mathbf{t}_{t+k}- \mathbf{R}_{t}^\top \mathbf{t}_{t}||_2}$ $\text{s.t.}~|K| = N$, $T_t(\mathbf{x})=\mathbf{R}_t\mathbf{x}+\mathbf{t}_t$ is the ego-motion from the world origin at time $t$, and $\mathcal{P}_t$ is 3D points captured at time $t$.
\vspace{2mm}

\noindent\textbf{Dynamic objects.} 
When we aggregate 3D points on dynamic objects in the world model, we should avoid unintended flying points occurring by object-wise motion. To alleviate the issue, we leverage temporally consecutive LiDAR frames, not the geometrically adjacent frames, and we consider trajectory information of the dynamic objects over time. In short, the sparse observations of dynamic objects across multiple frames are aggregated by applying inverse motions of each dynamic object and ego-motions.
\vspace{2mm}

\noindent\textbf{Label consistency and label propagation.}
After the world model construction, we examine the labeling consistency for all the aggregated 3D points in $\mathcal{P}_t^{world}$. This verification step is a safeguard to remove noisy points from various sources of errors, such as incorrect annotation and inaccurate ego-motion. To make consistent labels, we examine a set of 3D points assigned to a single voxel in the voxel grid (10cm). The majority voting determines a representative semantic label for each voxel, and we can get clean labels. Note that the majority label in a voxel can be propagated to the unlabeled points in the same voxel. This step helps assign pseudo labels to sparsely annotated datasets like nuScenes~\cite{nuscenes_lidarseg} that only provides dense annotations for keyframes selected at 2Hz.

\subsection{Creating a Range Map}
\label{subsec:sensor-agnostic LiDAR projection}

\noindent\textbf{Pose augmentation.}
Once we have a world model, the LiDAR pose is augmented by applying a rigid transformation $T_{aug}(\mathbf{x})=\textbf{R}_{aug}\mathbf{x}+\textbf{t}_{aug}$ to give variations of the LiDAR frames. In our experiments, random rotation along the z-axis, i.e., $\textbf{R}_{aug}=\textbf{R}_z(\theta_{yaw})$, and random translation are considered\footnote{While our method is capable of incorporating arbitrary rotations, it is worth noting that most public datasets utilize \emph{upright LiDARs}. Therefore, applying full rotations may result in an unintended severe domain gap.}. The yaw angle $\theta_{yaw}$ and the translation vector $\textbf{t}_{aug}$ are drawn from uniform distributions.
\vspace{2mm}

\noindent\textbf{Randomized LiDAR configurations.}
A LiDAR frame can be expressed as a range map, and the configuration of LiDAR is defined by the vertical field of view ($f_{up}$, $f_{down}$) and the resolution of the range map ($H$, $W$). In the case of cylindrical LiDARs, the projection $\Pi (\mathbf{x}) \rightarrow [u, v, r]^{\top}$ of the 3D points is calculated\footnote{We set the x-axis as the vehicle's forward direction, the y-axis as the left from the vehicle, and the z-axis as the top direction from the ground.} as follows~\cite{behley2018rss,domain_transfer_iros20}:

\begin{equation}
    \Pi(\begin{bmatrix} x \\ y \\ z \end{bmatrix}) 
    =
    \begin{bmatrix} \frac{1}{2}[1-(\arctan{\frac{y}{x}})/\pi]W \\ [1-(\arcsin{\frac{z}{r}}-f_{down})/f]H \\ ||\mathbf{x}||_2 \end{bmatrix}
    =
    \begin{bmatrix} u \\ v \\ r \end{bmatrix},
    \label{equ:projection}
\end{equation}
where $f=|f_{up}| + |f_{down}|$. With $\Pi(\cdot)$, we can project the world model $\mathbf{x}\in T_{aug}(\mathcal{P}_t^{world})$ using a given LiDAR configuration. Here, we \emph{randomize LiDAR configuration $(H,W,f_{up},f_{down})$} to augment LiDAR frames further. With this procedure, a range map of random LiDAR configuration can be rendered, and LiDAR patterns not observed in the training data can be provided. This step is shown to be very effective in our experiment.

The world models are constructed by aggregating LiDAR frames of different viewpoints, which can result in occlusion from a desired viewpoint. To filter out these points, we employ z-buffer-based raycasting~\cite{Z-buffer} that selects the nearest 3D points to the desired viewpoint. Therefore, we formulate the step for range image rendering as follows:
\begin{equation}
    Prj(\mathcal{P}_{t}) = \mathcal{Z}(\Pi(T_{aug}(\mathcal{P}_{t}^{world}))),
\end{equation}
where the $\mathcal{Z}$ means the z-buffer-based ray-casting.

\begin{figure}[t]
    \begin{center}
        \includegraphics[width=1\linewidth]{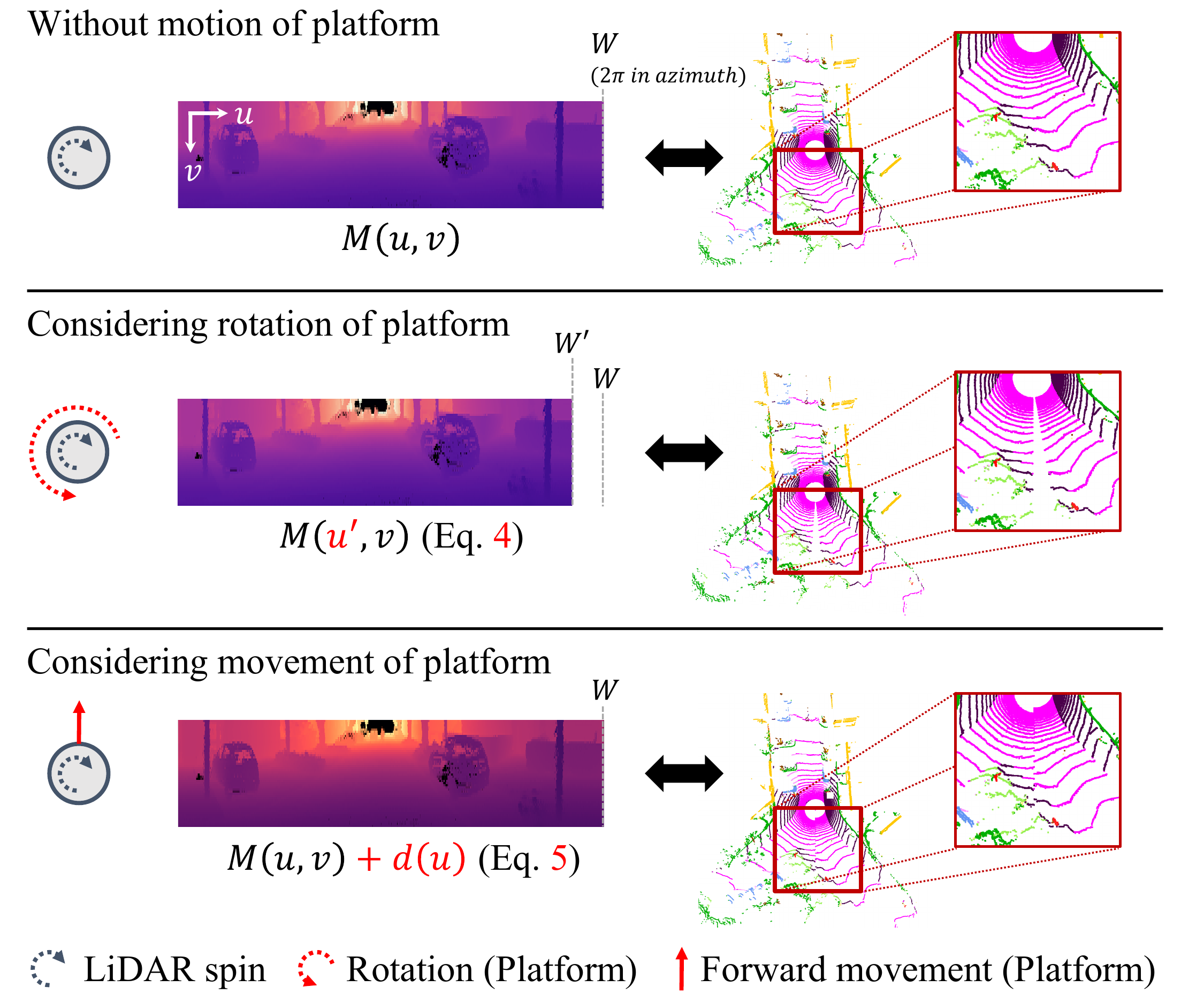}
    \end{center}
    \vspace{-10pt}
    
    \caption{
        Implementation of LiDAR distortion induced by entangled motion. Given movement/rotation velocities, distortion maps are generated and used to implement the motion distortion.         
    }
    \label{fig:distortion_and_occlusion}    
\end{figure}

\subsection{Adding Motion Distortion}
\label{subsec:Apply Situation-aware Transformation}

Cylindrical LiDARs have a spinning motion with a fixed rate for omnidirectional capture. Also, LiDARs are often mounted on a moving platform, such as a vehicle. The two entangled motions, i.e., movement of the vehicle and spinning motion of LiDAR, result in distortion on framed data. We observe such distortions from the real LiDAR frames (See the supplement). 

More specifically, the rotation of the platform affects the effective LiDAR angular velocity, resulting in a gap or overlap between starting and ending points of a single LiDAR frame shown in the middle of Fig.~\ref{fig:distortion_and_occlusion}. If the platform has a forward movement, as depicted at the bottom of Fig.~\ref{fig:distortion_and_occlusion}, the starting and ending points are not aligned because each 3D point has a different travel distance. Although this distortion could significantly change the coordinates of 3D points in a LiDAR frame, this phenomenon is rarely addressed in the literature.

We formulate the distortion with LiDAR spin angular velocity $\omega_0$, platform rotation angular velocity $\omega$, and platform forward movement velocity $V$ under constant velocity assumption. 
\begin{align}
    M(u', v) = M(\frac{(\omega_0+\omega)}{\omega_0}\cdot u, v),
    \label{equ:distortion_rotation} \\
    M(u, v) + d(u) = M(u, v) + V\cdot\frac{u}{\omega_0},
    \label{equ:distortion_movement}
\end{align}

where $M$ is a range map projection of a LiDAR frame.
The effective angular velocity is $\omega+\omega_0$ for distortion by rotation, which results in a resampling of each 3D point in the range map along $u$-axis, as shown in Eq.~\ref{equ:distortion_rotation}. The travel distance compensation due to the forward movement is given by Eq.~\ref{equ:distortion_movement}. These equations lead us to an efficient implementation of the \emph{distortion in the range map} by applying coordinate resampling and depth adjustment.

\subsection{Scene-level \& Sensor-level Mix}
\label{sec:sensor_level_mix}

PolarMix~\cite{Polarmix} shows a scene-level mix and demonstrates strong data augmentation performance. Inspired by this work, we propose an extended augmentation module that mixes frames of \emph{different scenes captured by different LiDARs}. As described in Fig.~\ref{fig:overview}, after rendering range maps with random LiDAR configurations, we mix the range maps using random azimuth angle ranges. The mixed range map is transformed to a 3D point cloud using the inverse of the projection model $\Pi^{-1}$. Generally, the diversity of training data in a single training step is proportional to the batch size. The mixing module helps data-driven approaches by providing diverse LiDAR patterns in a single batch, reducing efforts to keep a large batch size.

%% file: section/experiment.tex
\section{Experiment}
\label{sec:experiment}

We conduct a series of experiments to test the generalization ability over the sensor-bias issue in the domain adaptation setting, in which a different LiDAR sensor is used at test time. We compare our method with domain adaptation and data augmentation approaches (Sec.~\ref{sec:exp:domain_adaptation}). Next, we demonstrate the effectiveness of our method in training a sensor-unbiased model (Sec.~\ref{sec:exp:towards_unbiased_model}). Last, we perform an ablation study on each technical contribution (Sec.~\ref{sec:exp:ablation}).

\subsection{Implementation details}
\label{sec:implementation}

In our experiment, as described in Sec.~\ref{subsec:sensor-agnostic LiDAR projection}, the yaw angle in the rotation matrix is randomly sampled from a uniform distribution, $\theta_{yaw}\sim U(-\frac{\pi}{6}, \frac{\pi}{6})$, and each element in the translation vector is drawn from another uniform distribution, $\textbf{t}_{aug}=[x, y, z]^T$ where $x\sim U(-1, 1)$m, $y\sim U(-0.5, 0.5)$m, and $z\sim U(-0.1, 0.1)$m. In addition, we set random LiDAR configuration parameters that are described in Sec.~\ref{subsec:Apply Situation-aware Transformation} as follows: $H \in \{1024, 2048\}$px., $W\in[16, 128]$px., $f_{up}\in[0,\pi/12)$, $f_{down}\in[-\pi/6,0)$. Note that we sample $W$, $f_{up}$, and $f_{down}$ from certain ranges to render arbitrary configurations of LiDARs. The forward movement velocity $V$ is sampled from $U(0, 60)$~km/h, and the rotation angular velocity of the vehicle $\omega$ is sampled from $U(-\frac{\pi}{8}, \frac{\pi}{8})$. We mix two augmented LiDAR frames in our experiment. 

We implement every module with GPU primitives for speed gain. As a result, our method can be seamlessly plugged into the data loader in the training pipeline due to its efficiency. For example, our method integrated into the data loader of MinkNet42 network training adds just 3ms to render a new LiDAR frame.\footnote{We use a workstation equipped with AMD EPYC 7452 CPU and Nvidia GeForce RTX 3090 GPU.}

\subsection{Datasets}

\noindent\textbf{SemanticKITTI}~\cite{semantic_kitti} is a large-scale dataset for LiDAR semantic segmentation task built upon the popular KITTI Vision Odometry Benchmark~\cite{kitti}. It consists of 22 sequences with 19 annotated classes. The dataset was collected by a Velodyne HDL-64E that has 64 vertical beams for 26.9$\degree$ of vertical field of view (+2.0$\degree$ to -24.9$\degree$) corresponding to $64\times2048$ range map. Following the standard protocol, we use sequences 00 to 10 (19k frames) for training except sequence 08 (4k frames), reserved for a validation set. Since SemanticKITTI does not provide the 3D bounding boxes, we treat the dynamic objects as a part of the static scene when constructing the world models described in Sec.~\ref{sec:construct_a_world_model}. 
\vspace{2mm}

\noindent\textbf{nuScenes-lidarseg}~\cite{nuscenes_lidarseg} is another large dataset providing 1,000 driving scenes (850 for training and validation, 150 for testing), including per-point annotation for 16 categories. However, as only the keyframes sampled at 2Hz are annotated, the label propagation scheme described in Sec.~\ref{sec:construct_a_world_model} is applied. This dataset was captured with a Velodyne HDL-32E, providing 32 vertical beams for 41.33$\degree$ of vertical field of view (+10.67$\degree$ to -30.67$\degree$), resulting in $32\times2048$ range map. We consider each motion of dynamic objects in constructing world models using the given 3D bounding box trajectory information.
\vspace{2mm}

\noindent\textbf{Label Mapping.} 
Since the annotated classes in Semantic KITTI\cite{semantic_kitti} and nuScene\cite{nuscenes_lidarseg} differ, we evaluate only ten overlapping categories in our experiments: \{Car, Bicycle, Motorcycle, Truck, Other vehicles, Pedestrian, Drivable surface, Sidewalk, Terrain, and Vegetation\} as suggested by \cite{complete_and_label_cvpr21}. We use mean Intersection-over-Union(mIoU) as our evaluation metric.

\begin{table}[t]
    \newcommand{\minus}[1]{{\textcolor{purple}{$\downarrow$ #1}}}
    \newcommand{\plus}[1]{{\textcolor{teal}{$\uparrow$ #1}}}
    \newcommand{\ccgreen}{\cellcolor{yellow!10}}
    
    \newcommand{\xmark}{\textcolor{teal}{\ding{55}}}%
    \newcommand{\cmark}{\textcolor{purple}{\ding{51}}}%

    \newcolumntype{L}{>{\hspace{0.7em}}l}
    \newcolumntype{R}{>{\hspace{0.6em}}r}

    \centering
    \resizebox{1\columnwidth}{!}{
        \setlength{\tabcolsep}{2pt}
        \begin{tabular}{c|L|R@{\hspace{0.5em}}lR@{\hspace{0.5em}}l}    
        \multicolumn{1}{c}{}&\multicolumn{1}{c}{}&\multicolumn{4}{r}{Unit: mIoU (Rel.\%)}\\
        \toprule
        \multirow{2}{*}{\begin{tabular}[c]{@{}c@{}}Backbone\\(\# of params)\end{tabular}}
        &\multicolumn{1}{c|}{\multirow{2}{*}{\begin{tabular}[c]{@{}c@{}}Methods\end{tabular}}}
        &\multicolumn{4}{c}{Source $\rightarrow$ Target} \\
        \cline{3-6} 
        \multicolumn{1}{c|}{} & \multicolumn{1}{c|}{} &\multicolumn{2}{c}{$K(64) \rightarrow N(32)$} & \multicolumn{2}{c}{$N(32) \rightarrow K(64)$} \\         
        \cline{1-6}
        \multicolumn{1}{c|}{\multirow{7}{*}{\shortstack{Complete\\ \&Label~\cite{complete_and_label_cvpr21}\\(8.39M)}}} 
        & Baseline & \multicolumn{2}{c}{27.9} & \multicolumn{2}{c}{23.5} \\
        \cline{2-6}
        & FeaDA~\cite{FeaDA}  & 27.2 & (\minus{2.5}) & 21.4 & (\minus{8.9}) \\
        & OutDA~\cite{OutDA}  & 26.5 & (\minus{5.0}) & 22.7 & (\minus{3.4}) \\
        & SWD~\cite{SWD}      & 27.7 & (\minus{0.7}) & 24.5 & (\minus{4.3}) \\
        & 3DGCA~\cite{3DGCA}  & 27.4 & (\minus{1.8}) & 23.9 & (\plus{1.7}) \\
        &\Centerstack{C\&L~\cite{complete_and_label_cvpr21}} & 31.6 & (\textbf{\plus{13.3}}) & 33.7 & (\plus{43.4})\\        
        &\ccgreen&\ccgreen&\ccgreen&\ccgreen&\ccgreen\\
        &\multirow{-2}{*}{\ccgreen \shortstack[l]{Baseline\\+ \OursAcronym{}}} & \multirow{-2}{*}{\ccgreen \textbf{39.2}} & \multirow{-2}{*}{\ccgreen(\textbf{\plus{40.5}})}& \multirow{-2}{*}{\ccgreen \textbf{37.9}} & \multirow{-2}{*}{\ccgreen(\textbf{\plus{61.3}})} \\
        \cmidrule[0.8pt]{1-6}
        \multicolumn{6}{c}{(a) Comparison with unsupervised domain adaptation approaches} \\
        \multicolumn{6}{c}{} \\
        
        \multicolumn{1}{c}{}&\multicolumn{1}{c}{}&\multicolumn{4}{r}{Unit: mIoU (Rel.\%)}\\
        \toprule
        \multirow{2}{*}{\begin{tabular}[c]{@{}c@{}}Backbone\\(\# of params)\end{tabular}}
        &\multicolumn{1}{c|}{\multirow{2}{*}{\begin{tabular}[c]{@{}c@{}}Methods\end{tabular}}}
        &\multicolumn{4}{c}{Source $\rightarrow$ Target} \\
        \cline{3-6}
        \multicolumn{1}{c|}{} & \multicolumn{1}{c|}{} &\multicolumn{2}{c}{$K(64) \rightarrow N(32)$} & \multicolumn{2}{c}{$N(32) \rightarrow K(64)$} \\         
        \cline{1-6}
        \multicolumn{1}{c|}{\multirow{6}{*}{\Centerstack{MinkNet42~\cite{Minkowski}\\(37.8M)}}} 
        & Baseline & \multicolumn{2}{c}{37.8} & \multicolumn{2}{c}{36.1} \\
        \cline{2-6}
        & CutMix~\cite{Cutmix} & 37.1 & (\minus{1.9}) & 37.6 & (\plus{4.2}) \\
        & Copy-Paste~\cite{Copy-Paste} & 38.5 & (\plus{1.9}) & 41.1 & (\plus{13.9}) \\
        & Mix3D~\cite{Mix3d} & 43.1 & (\plus{14.0}) & 44.7 & (\plus{23.8}) \\
        & PolarMix~\cite{Polarmix} & 45.8 & (\plus{21.2}) & 39.1 & (\plus{8.3}) \\        
        &\ccgreen&\ccgreen&\ccgreen&\ccgreen&\ccgreen\\
        & \multirow{-2}{*}{\ccgreen \shortstack[l]{Baseline\\+ \OursAcronym{}}}  & \multirow{-2}{*}{\ccgreen \textbf{45.9}} & \multirow{-2}{*}{\ccgreen(\textbf{\plus{21.4}})}& \multirow{-2}{*}{\ccgreen \textbf{48.3}} & \multirow{-2}{*}{\ccgreen(\textbf{\plus{33.8}})}\\
        \cmidrule[0.8pt]{1-6}
        \multicolumn{6}{c}{(b) Comparison with data augmentation approaches}
        \end{tabular}
    }    
    
    \caption{                
        An experiment with domain adaptation settings. We train networks with SemanticKITTI (64 ch.)~\cite{kitti} and test with nuScenes (32 ch.)~\cite{nuscenes_lidarseg} (K$\rightarrow$N) and vice versa (N$\rightarrow$K). Baseline with the proposed \OursAcronym{} is more effective than state-of-the-art Domain Adaptation and Data Augmentation approaches.         
    }    
    \label{tab:Domain_adaptation}
\end{table}

\subsection{Results}
\label{sec:exp:domain_adaptation}

We compare our method with unsupervised domain adaptation and domain mapping methods. All the methods are trained on SemanticKITTI\cite{semantic_kitti} and evaluated on nuScenes\cite{nuscenes_lidarseg} (K$\rightarrow$N) or vice versa (N$\rightarrow$K). Note that our approach \emph{does not utilize the target dataset nor utilize the target LiDAR sensor information}. As stated in Sec.~\ref{subsec:sensor-agnostic LiDAR projection}, our approach is trained with randomized LiDAR configurations described in Sec.~\ref{sec:implementation} for this experiment.
\vspace{2mm}

\noindent\textbf{Unsupervised Domain Adaptation.} 
As shown in Table~\ref{tab:Domain_adaptation}-(a), our method shows consistent improvement by a large margin over the state-of-the-art methods in both adaptation settings (K$\rightarrow$N and N$\rightarrow$K). In the N$\rightarrow$K setting, for example, even though the model is trained on sparse data (32-ch) than the target domain (64-ch), our augmentation method provides more density-varied examples than what is available in the source domain, which helps improve the learning of sensor-agnostic representations.

Adversarial domain alignment methods, FeaDA~\cite{FeaDA}, OutDA~\cite{OutDA}, SWD~\cite{SWD}, and 3DGCA~\cite{3DGCA}, show similar performance with the baseline and reveal the limitation in learning sensor-unbiased representations.\footnote{As reported in~\cite{complete_and_label_cvpr21}, previous domain adaptation methods, namely FeaDA, OutDA, SWD, and 3DGCA, are ineffective in handling 3D LiDAR data. For more details, please refer to Sec. 4.2 in ~\cite{complete_and_label_cvpr21}.} Compared with C\&L~\cite{complete_and_label_cvpr21} that requires additional back-and-forth mapping to the canonical domain at the test time, our augmentation method is just applied at the training time, and it does not add additional computational burdens at the test time. 

\begin{table}[bt]

    \newcommand{\ccgreen}{\cellcolor{yellow!10}}
    \newcommand{\minus}[1]{{\textcolor{purple}{$\downarrow$ #1}}}
    \newcommand{\plus}[1]{{\textcolor{teal}{$\uparrow$ #1}}}
    
    \newcolumntype{L}{>{\hspace{0.7em}}l}
    \newcolumntype{R}{>{\hspace{1em}}r}
    
    \centering
    
    \resizebox{0.95\columnwidth}{!}{
        \setlength{\tabcolsep}{2pt}
        \begin{tabular}{L|c|R@{\hspace{0.1em}}L@{\hspace{0.3em}}}
        \multicolumn{4}{r}{Unit: mIoU (Rel.\%)}\\
        \toprule
        \multirow{2}{*}{Method} & \multirow{2}{*}{Retraining} & \multicolumn{2}{c}{Source $\rightarrow$ Target} \\
        \cline{3-4}
        & & \multicolumn{2}{c}{$K(64) \rightarrow N_{0103}(32)$} \\
        \midrule    
        CP\cite{domain_transfer_iros20} & \textcolor{teal}{No} & \multicolumn{2}{c}{28.8}\\
        MB\cite{domain_transfer_iros20} & \textcolor{teal}{No} & 30.0 & (\plus{4.2}) \\
        MB+GCA\cite{domain_transfer_iros20} & \textcolor{purple}{Required} & 32.6 & (\plus{13.2}) \\
        CP+GCA\cite{domain_transfer_iros20} & \textcolor{purple}{Required} & 35.9 & (\plus{24.7})\\    
        BonnetalPS+AdaptLPS\cite{bevsic2021unsupervised} & \textcolor{purple}{Required} & 37.5 & (\plus{30.2})\\
        EfficientLPS+AdaptLPS\cite{bevsic2021unsupervised} & \textcolor{purple}{Required} & 38.5 & (\plus{33.7})\\    
        \ccgreen MinkNet42 + \OursAcronym{} & \ccgreen \textcolor{teal}{No} & \ccgreen \textbf{52.4} & \ccgreen (\plus{81.9}) \\
        \bottomrule
        \end{tabular}
    }
        
    \caption{
        Comparison with domain mapping approaches. We follow the evaluation protocol used in \cite{domain_transfer_iros20}, i.e., trained on SemanticKITTI (K) and evaluated on a subset of the nuScene (N-0103), to ensure a fair comparison. Note that our approach does not require retraining and achieves performance improvement.        
    }
    \label{tab:domain_mapping}

\end{table}
\vspace{2mm}

\noindent\textbf{Domain Mapping.}
Domain Mapping methods~\cite{bevsic2021unsupervised,domain_transfer_iros20} try to convert the source domain data to target domain data as closely as possible, so they are required to access the target domain data. Some approaches, such as GCA~\cite{domain_transfer_iros20} and AdaptLPS~\cite{bevsic2021unsupervised}, as shown in Table~\ref{tab:domain_mapping}, even require retraining networks. On the other hand, as discussed in Sec.~\ref{sec:related work}, our method is an effective instant domain augmentation approach, which provides diverse LiDAR patterns beyond the target domain patterns during training, so it is a good alternative to the domain mapping approaches. Table~\ref{tab:domain_mapping} shows the comparison result. We follow the same evaluation protocol used in \cite{domain_transfer_iros20} for a fair comparison, and our model shows superior performance over the state-of-the-art (BPS+AdaptLPS) by a large margin (38.5 vs. 52.4 mIOU), without re-training process required by the other approaches.

\begin{table*}[bt]
    \setlength{\tabcolsep}{3pt}
    \newcommand{\cboxgray}[1]{{\setlength{\fboxsep}{1pt}\colorbox{gray!20}{\setlength{\fboxsep}{1pt}#1}}}
    \newcommand{\cboxblue}[1]{{\setlength{\fboxsep}{1pt}\colorbox{blue!10}{\setlength{\fboxsep}{1pt}#1}}}
    
    \newcommand{\ccblue}{\cellcolor{blue!10}}
    \newcommand{\ccgray}{\cellcolor{gray!20}}
    \newcommand{\ccgreen}{\cellcolor{yellow!20}}
    
    \newcommand{\minus}[1]{{\textcolor{purple}{$\downarrow$ #1}}}
    \newcommand{\plus}[1]{{\textcolor{teal}{$\uparrow$ #1}}}
    \newcommand{\boldminus}[1]{{\textcolor{purple}{\boldsymbol{$\downarrow$} #1}}}
    \newcommand{\boldplus}[1]{{\textcolor{teal}{\boldsymbol{$\uparrow$} #1}}}
    \centering
    
    \resizebox{1.995\columnwidth}{!}{
        \begin{tabular}{l|c|cc|cc|cc|cc|cc|c}
            \multicolumn{1}{c}{}&\multicolumn{1}{c}{}&\multicolumn{1}{c}{}&\multicolumn{1}{c}{}&\multicolumn{1}{c}{}&\multicolumn{1}{c}{}&\multicolumn{1}{c}{}&\multicolumn{1}{c}{}&\multicolumn{1}{c}{}&\multicolumn{1}{c}{}&\multicolumn{3}{r}{Unit: mIoU (Rel.\%)}\\
            \toprule
            \multirow{2}{*}{Backbone} &
            \multicolumn{1}{c|}{\multirow{2}{*}{\shortstack{Training data}}} &
            \multicolumn{10}{c|}{Testing data} &
            \multirow{2}{*}{\shortstack{Avg. rank}} \\
            \cline{3-12}
            & \multicolumn{1}{c|}{} 
            & \multicolumn{2}{c|}{V64} 
             & \multicolumn{2}{c}{V32} 
            & \multicolumn{2}{c}{V16} 
            & \multicolumn{2}{c}{O64} 
            & \multicolumn{2}{c|}{O128} & \\
            \midrule
            KPConv\cite{KPConv} & \multicolumn{1}{c|}{V64} & \multicolumn{2}{c|}{\ccgray 54.70} & 0.02& (\minus{99.95}) & 0.01& (\minus{99.98}) & 0.01& (\minus{99.98}) & 0.01& (\minus{99.98}) & 6.2 \\    
            \cmidrule[0.8pt]{1-13}
            \multirow{7}{*}{\shortstack{MinkNet42~\cite{Minkowski}}}& \multicolumn{1}{c|}{V64} & \multicolumn{2}{c|}{\ccgray \textbf{62.80}} & 24.32& (\minus{43.65}) & 13.77& (\minus{59.64})& 25.80& (\minus{36.76}) & 24.05& (\minus{46.14})& 4.8 \\
            \cmidrule[0.8pt]{2-13}
            & V32  & 45.59& (\minus{27.40})& \multicolumn{2}{c|}{\ccgray 43.16} & 25.35& (\minus{25.70})& 29.05& (\minus{28.80})& 29.28& (\minus{34.42})& 4.0 \\
            & V16  & 33.70& (\minus{46.33})& 29.66& (\minus{31.27})& \multicolumn{2}{c|}{\ccgray \textbf{34.12}}& 34.07& (\minus{16.50})& 31.48& (\minus{29.50})& 4.0 \\
            & O64  & 43.01& (\minus{31.51}) & 39.13& (\minus{9.34}) & 27.96& (\minus{18.05}) & \multicolumn{2}{c|}{\ccgray \uline{40.80}} & 43.08& (\minus{3.516}) & 3.2 \\
            & O128 & 42.25& (\minus{32.72}) & 27.41& (\minus{36.49}) & 10.54& (\minus{69.11}) & 37.81& (\minus{7.33}) & \multicolumn{2}{c|}{\ccgray \uline{44.65}} & 4.4 \\
            & \ccgreen \OursAcronym{} (Rand) & \ccgreen \uline{61.51}& \ccgreen (\minus{2.05}) & \ccgreen \textbf{44.73}& \ccgreen (\textbf{\plus{3.64}}) & \ccgreen \uline{33.38} & \ccgreen (\minus{2.17}) & \ccgreen \textbf{46.54} & \ccgreen (\textbf{\plus{14.07}}) & \ccgreen \textbf{48.34} & \ccgreen (\textbf{\plus{8.26}}) & \ccgreen \textbf{1.4} \\
            \bottomrule
        \end{tabular}
    }

    \caption{
        Experiment on the sensor-bias issue of LiDAR semantic segmentation models. We tested KPConv~\cite{KPConv} and MinkowskiNet~\cite{Minkowski} models on SemanticKITTI~\cite{semantic_kitti} by generating multiple LiDAR configurations using the proposed \OursAcronym{}. `\OursAcronym{} (Rand)' denotes that the model is trained with our final method, i.e., with pose-augmentation, randomized LiDAR config., random distortion, and scene-level \& sensor-level mixing. For each test configuration, \textbf{the best} and \uline{the second-best} performances are highlighted, the reference cases (the LiDAR scans come from the same LiDAR configurations are used at the training and the testing time) are \cboxgray{colored in gray}.         
    }    
    
    \label{tab:simulated_data_experiment}    
\end{table*}
\vspace{2mm}

\noindent\textbf{Data Augmentation.}
Although the 3D augmentation methods~\cite{Cutmix, Copy-Paste, Mix3d, Polarmix} have shown their effectiveness in learning a good representation on a \emph{single domain}, it is rarely studied to show the effectiveness of domain adaptation settings particularly caused by sensor discrepancy. We experiment to see whether the existing LiDAR augmentation methods and our approach are good at domain adaptation. 

As shown in Table~\ref{tab:Domain_adaptation}-(b), interestingly, the 3D augmentation methods~\cite{Cutmix, Copy-Paste, Mix3d, Polarmix} are helpful in domain adaptation settings, even though they are not designed for adapting to an unseen domain. In particular, Mix3D~\cite{Mix3d} shows impressive improvements (37.8 $\rightarrow$ 43.1 and 36.1 $\rightarrow$ 44.7) by simply aggregating two 3D scenes. However, we speculate that the point cloud aggregation by Mix3D can induce unusual local structures (e.g., two cars overlapped perpendicularly), which may result in a suboptimal model.

Furthermore, PolarMix~\cite{Polarmix}, works well in the K$\rightarrow$N setting (37.8 $\rightarrow$ 45.8). Our conjecture for the success of PolarMix in the K$\rightarrow$N setting is that it can provide patterns of sparse 3D points similar to those found in nuScenes (32-ch) when faraway portions of a KITTI frame (64-ch) are selected for mixing. However, if the source domain does not provide enough diversity as in the opposite N$\rightarrow$K scenario, PolarMix shows reduced improvement (36.1 $\rightarrow$ 39.1). This result shows that learning a rich sensor-agnostic representation is challenging. Our method aims to reduce the domain gap induced by sensor discrepancy by explicitly rendering various LiDAR patterns. As a result, our method achieves superior performances in both K$\rightarrow$N and N$\rightarrow$K settings by a large margin (37.8 $\rightarrow$ 45.9 and 36.1 $\rightarrow$ 48.3).

\subsection{Towards Sensor-agnostic Model}
\label{sec:exp:towards_unbiased_model}

Our method encourages models to learn a sensor-agnostic representation, and no data from the target domain is required during training. In this experiment, we discuss the effectiveness of our approach in training a model unbiased to any LiDAR configuration. This experiment is challenging to conduct because there is no real-world dataset captured by different kinds of LiDARs at once\footnote{Carballo~\etal~\cite{carballo2020libre} proposed a multi-lidar dataset, but the dataset is not released to the public domain.}. Therefore, it is not straightforward to configure a dataset of the same scene captured with different LiDARs. 

To proceed with this experiment, we use the proposed~\OursAcronym{} to create LiDAR frames of various LiDAR configurations from the SemanticKITTI dataset. These frames of specific LiDAR configurations were then mix-and-matched for training and testing datasets. Specifically, we create the frames of 16-, 32-, and 64-ch Velodyne LiDARs~\cite{velodyne2007} (denoted by V16, V32, and V64) and the frames of 64- and 128-ch Ouster LiDARs~\cite{ouster2018} (denoted by O64 and O128) based on the manufacturer-provided LiDAR specification\footnote{More detailed configurations are described in the supplement.}. We tested KPConv~\cite{KPConv} and MinkowskiNet~\cite{Minkowski}, which are representative methods in point- and voxel-based approaches, respectively.

In Table~\ref{tab:simulated_data_experiment}, we present evaluation results on various LiDAR patterns (columns) of a model trained on a specific LiDAR pattern (row). As shown in rows 1-2, the models show severe performance drops if the LiDAR got changed at test time. For instance, a MinkNet42~\cite{Minkowski} model trained on V64 LiDAR pattern in the second row achieves 62.80 mIoU if tested on the same LiDAR. However, the model shows significant performance drops if evaluated on different LiDAR patterns (24.32 mIoU on V32, 13.77 mIoU on V16, etc.). Especially, KPConv~\cite{KPConv} fails under this sensor-discrepancy scenario while MinkowskiNet~\cite{Minkowski} model is less affected. We speculate that the U-Net style of architectural design makes it resilient to variations of the geometric patterns of 3D points. Therefore, we choose MinkowskiNet~\cite{Minkowski} as the backbone model for the rest of the experiment.

We also trained MinkowskiNet~\cite{Minkowski} models on the other LiDAR configurations, such as V32, V16, O64, and O128, shown in rows 3-6 of Table~\ref{tab:simulated_data_experiment}. As expected in the sensor-discrepancy evaluation scenarios, the best performances are achieved when the same LiDAR is applied at test time (colored in gray). Otherwise, the performances fluctuate a lot. This result indicates that a data-driven model tends to be biased towards a specific LiDAR configuration of the training data, which could be a hurdle in deploying them to real-world applications.

As a remedy for the sensor-bias issue, we propose to train models with the proposed \OursAcronym{} using \emph{randomized LiDAR configurations}, shown in row 7 of Table~\ref{tab:simulated_data_experiment}. Our model gets to learn sensor-agnostic representations since \OursAcronym{} provides various LiDAR patterns with realistic distortions. Given the \emph{no free lunch theorem}~\cite{shalev2014understanding}, \OursAcronym{} (Rand) does not beat all the test settings, especially when the LiDAR configuration used for the training and testing is the same. Our model, however, achieves the highest generalization ability across the diverse LiDAR configurations we tested, measured by average rank in the last column of Table~\ref{tab:simulated_data_experiment}. This result shows that the proposed method helps alleviate the sensor bias.

\subsection{Ablation Study}
\label{sec:exp:ablation}

\begin{table}[t]
    \setlength{\tabcolsep}{3pt}
    \newcommand{\ccgreen}{\cellcolor{yellow!10}}    
    \newcommand{\minus}[1]{{\textcolor{purple}{$\downarrow$ #1}}}
    \newcommand{\plus}[1]{{\textcolor{teal}{$\uparrow$ #1}}}
    \centering
    
    \resizebox{1.0\columnwidth}{!}{
        \begin{tabular}{cccc|c}
        &&&\multicolumn{2}{r}{Unit: mIoU (Rel.\%)}\\
        \toprule
        Training data & Pose-Aug & Distortion & S\&S Mix & \multirow{2}{*}{$K(64) \rightarrow N(32)$} \\
        (Sec~\ref{subsec:sensor-agnostic LiDAR projection})
        & (Sec~\ref{subsec:sensor-agnostic LiDAR projection}) 
        & (Sec~\ref{subsec:Apply Situation-aware Transformation})
        & (Sec~\ref{sec:sensor_level_mix}) & \\
        \midrule
        K(64) &&&& 36.57$\pm$0.56 \\    
        \midrule
        K(32) &&&& 37.05$\pm$1.15 (\plus{1.31}) \\
        Random &&&& 40.05$\pm$1.03 (\plus{9.51}) \\ 
        Random &\checkmark &&& 42.70$\pm$0.91 (\plus{16.8}) \\
        Random &\checkmark &\checkmark && 43.04$\pm$0.56 (\plus{17.7}) \\    
        \ccgreen Random &\ccgreen \checkmark &\ccgreen \checkmark &\ccgreen \checkmark & \ccgreen \textbf{44.98$\pm$1.42} (\plus{\textbf{23.0}}) \\
        \bottomrule
        \end{tabular}
    }
    
    \vspace{-2mm}
    \caption{
        Ablation study on the proposed augmentation modules. The results are averaged over three runs.
    }    
    \label{tab:ablation}
    \vspace{-2mm}
\end{table}

\begin{table}[t]
    \newcommand{\minus}[1]{{\textcolor{purple}{$\downarrow$ #1}}}
    \newcommand{\plus}[1]{{\textcolor{teal}{$\uparrow$ #1}}}
    \newcommand{\ccgreen}{\cellcolor{yellow!10}}
    \newcommand{\xmark}{\textcolor{teal}{\ding{55}}}%
    \newcommand{\cmark}{\textcolor{purple}{\ding{51}}}%

    \newcolumntype{L}{>{\hspace{0.7em}}l}
    \newcolumntype{R}{>{\hspace{0.6em}}r}

    \vspace{-2mm}
    \centering
    \resizebox{1.0\columnwidth}{!}{
        \setlength{\tabcolsep}{2pt}
        \begin{tabular}{c|L|R@{\hspace{0.5em}}lR@{\hspace{0.5em}}l}    
            \multicolumn{1}{c}{}&\multicolumn{1}{c}{}&\multicolumn{4}{r}{Unit: mIoU (Rel.\%)}\\
            \toprule
            \multicolumn{1}{c|}{\multirow{2}{*}{\begin{tabular}[c]{@{}c@{}}Backbone\\(\# of params)\end{tabular}}}
            &\multicolumn{1}{c|}{\multirow{2}{*}{\begin{tabular}[c]{@{}c@{}}Methods\end{tabular}}}
            &\multicolumn{4}{c}{Source $\rightarrow$ Target} \\
            \cline{3-6} 
            \multicolumn{1}{c|}{} & \multicolumn{1}{c|}{} &\multicolumn{2}{c}{$K(64) \rightarrow N(32)$} & \multicolumn{2}{c}{$N(32) \rightarrow K(64)$} \\                 
            \cmidrule[0.8pt]{1-6}
            \multicolumn{1}{c|}{\multirow{3}{*}{\Centerstack{SPVCNN~\cite{SPVNAS_ECCV20}\\(37.9M)}}} 
            & Baseline & \multicolumn{2}{c}{43.4} & \multicolumn{2}{c}{41.9} \\
            &\ccgreen&\ccgreen&\ccgreen&\ccgreen&\ccgreen\\
            &\multirow{-2}{*}{\ccgreen\shortstack[l]{Baseline\\+ \OursAcronym{}}} & \multirow{-2}{*}{\ccgreen \textbf{51.7}} & \multirow{-2}{*}{\ccgreen(\textbf{\plus{19.1}})}& \multirow{-2}{*}{\ccgreen \textbf{51.2}} & \multirow{-2}{*}{\ccgreen(\textbf{\plus{22.2}})}\\
            \bottomrule
        \end{tabular}
    }

    \vspace{-2mm}
    \caption{                
        Training with SemanticKITTI~\cite{kitti} and testing with nuScenes-lidarseg~\cite{nuscenes_lidarseg} (K$\rightarrow$N). The vice versa is denoted as (N$\rightarrow$K). We utilize state-of-the-art NAS-based 3D neural network architecture (SPVCNN~\cite{SPVNAS_ECCV20}) for this experiment.
    }    
    \label{fig:spvcnn_experiment}    
\end{table}

We perform an ablation study on the impact of each proposed contribution. In this experiment, we train MinkNet42 models~\cite{Minkowski} on SemanticKITTI (64 ch.)~\cite{semantic_kitti} and test them on nuScenes (32 ch.)~\cite{nuscenes_lidarseg}, i.e., $K(64)\rightarrow N(32)$ scenario.
\vspace{2mm}

\noindent\textbf{Training with randomized LiDARs.}
We compare models trained on three types of LiDAR patterns: (1) the original SemanticKITTI dataset, denoted as K(64). (2) 32 ch. LiDAR, created by \OursAcronym{} using SemanticKITTI dataset and target nuScene LiDAR specification, denoted as K(32). (3) LiDAR scans created by \OursAcronym{} using randomized configurations, denoted as Random. As shown in rows 1-3 of Table~\ref{tab:ablation}, K(32) shows improvement over K(64) because K(32) resembles the data in the target domain, denoted as N(32). Random data provides examples with abundant patterns to help learn a better representation, achieving additional performance gain (36.57 $\rightarrow$ 40.05).
\vspace{2mm}

\noindent\textbf{Pose augmentation} is another effective source of providing a diversity of the scan patterns. 
As shown in Table~\ref{tab:ablation}, adding pose augmentation to the Random LiDAR inputs leads to extra improvement (40.05 $\rightarrow$ 42.70).
\vspace{2mm}

\noindent\textbf{Distortion induced by entangled motion.}
In the real world, LiDAR data have distortions due to the entangled motion of the vehicle and LiDAR. Our distortion module implements various degrees of distortion from randomly selected forward and angular velocities of the vehicle. As this module enhances the realism of LiDAR frames, we achieve another enhancement (42.70 $\rightarrow$ 43.04), shown in row 5 of Table~\ref{tab:ablation}.
\vspace{2mm}

\noindent\textbf{Scene-level \& Sensor-level mix} module provides extra diversity to a single LiDAR frame by swapping scenes captured with different LiDAR configurations. Eventually, the final model shown in row 6 of Table~\ref{tab:ablation} learns a better representation from the rich LiDAR frames (43.04 $\rightarrow$ 44.98).
\vspace{2mm}

\noindent\textbf{NAS-based backbone.} We additionally validate that our approach can be applied to an advanced 3D neural network, SPVCNN~\cite{SPVNAS_ECCV20} that was found by extensive neural architecture search (NAS). We use the same experimental setting used in Sec.~\ref{sec:exp:domain_adaptation}, and we utilize the pre-trained network provided by the authors. After fine-tuning the network with the proposed~\OursAcronym{}, we observe the prediction accuracy on the unseen target domain significantly enhanced in both $K(64)\rightarrow N(32)$ and $N(32)\rightarrow K(64)$ scenarios.

%% file: section/conclusion.tex
\section{Conclusion}
\label{sec:conclusion}
This paper proposes a new LiDAR augmentation method to remedy the sensor-bias issue in LiDAR semantic segmentation models. Our method efficiently transforms real-world LiDAR data to another LiDAR domain having the desired configuration. Due to its efficiency, our method can be deployed as an online data augmentation module in the learning frameworks, which leads us to call our method instant domain augmentation. Our method does not require access to any target data, so it encourages models to learn a sensor-agnostic representation by providing random LiDAR configurations of data. Extensive experiments show that training with our method significantly improves the LiDAR semantic segmentation performance in the unseen datasets collected by a different LiDAR.
\vspace{2mm}

\noindent\textbf{Limitation and future work.}
Our method requires accurate 6-DoF ego-motions to construct the world models, but it could be estimated by off-the-shelf LiDAR SLAM method~\cite{bai2022fasterlio}. Our method is crafted for a cylindrical LiDAR, the most common type utilized in most of the existing public datasets. However, our trivial extension to a more complex setting, i.e., two LiDAR settings consisting of a solid-state LiDAR and a cylindrical LiDAR, shows a promising result (see Section \textcolor{red}{E} in the supplement). Since our method can be used for a generic LiDAR domain augmentation, our future work is to apply our method to other 3D perception tasks, such as object detection or instance semantic segmentation.